%% file: main.tex
\documentclass[10pt,twocolumn,letterpaper]{article}

\usepackage{wacv}  %

\input{preamble}
\definecolor{wacvblue}{rgb}{0.21,0.49,0.74}
\usepackage{xr-hyper}
\usepackage[breaklinks,colorlinks,allcolors=wacvblue]{hyperref}
\makeatletter
\AtBeginDocument{{\let\bibcite\@gobbletwo \externaldocument{suppx}[supp.pdf]}}
\makeatother

\def\wacvPaperID{1140} %
\def\confName{WACV}
\def\confYear{2027}

\title{Auditing Grasp Feasibility Predictors with Planner-Verified Counterfactuals}

\author{Sajid Ansari\\
IIIT Hyderabad, India
\and
Antony Thomas\\
IIIT Hyderabad, India
\and
Girish Varma\\
IIIT Hyderabad, India
}

\begin{document}
\maketitle
\input{sec/0_abstract}    
\input{sec/1_intro}

\input{sec/2_related}
\input{sec/3_setup}
\input{sec/4_results}
\input{sec/5_conclusion}

\section*{Acknowledgments}
This work was supported by the Anusandhan National Research Foundation (ANRF), Government of India, under the Advanced Research Grant (ARG) scheme [grant number ANRF/ARG/2025/009038/ENS].

{
    \small
    \bibliographystyle{ieeenat_fullname}
    \bibliography{main}
}

\end{document}


\maketitlesupplementary

\setcounter{figure}{8}
\setcounter{table}{6}
\input{sec/6_supp}

{
    \small
    \bibliographystyle{ieeenat_fullname}
    \bibliography{main}
}

%% file: sec/0_abstract.tex
\begin{abstract}
Grasp-feasibility predictors report near-perfect accuracy on Task and Motion Planning benchmarks, but high accuracy on these benchmarks does not verify that a predictor uses the collision geometry that the motion planner checks. We propose a planner-verified counterfactual (CF) test: holding the grasp pose fixed, we insert a solid obstacle into the approach corridor, add its points to the model input, and let a planner re-derive the label, which flips from feasible to infeasible. Across eight predictors spanning point clouds, voxels, depth, and object tokens, CF-pair AUROC falls from high natural AUROC ($0.920$--$0.996$) to chance ($0.470$--$0.504$), and the flip holds across four planners, two arms, and a real-object obstacle. Predictors fail at different resolutions, from a bounding box to a $12.5$\,cm voxel. In the point-cloud models we dissect, the obstacle either moves the representation in a direction unrelated to whether it blocks the grasp or is not registered at all. A clearance-aware read of the corridor points, trained on the unmodified benchmark with no counterfactual scenes, reaches $0.983$ CF-pair AUROC, but it remains bound to its trained obstacle placements; a placement-invariant clearance representation is still open. We release the counterfactual labels, metrics, datasheet, and relabeling pipeline at \url{https://github.com/codewithsajid/graspfc-cf}.
\end{abstract}

%% file: sec/1_intro.tex
\section{Introduction}
\label{sec:intro}

\begin{figure}[t]
\centering
\includegraphics[width=\linewidth]{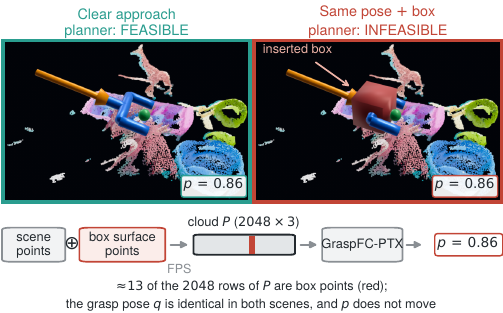}
\caption{The counterfactual test. (\textbf{left}) A feasible grasp. (\textbf{right}) The same grasp becomes infeasible once a cube (red) is inserted in its approach path; points sampled from the cube's surface are added to the point-cloud input. The classifier (GraspFC-PTX) still predicts $p(\text{feasible}) = 0.86$ in both scenes. Over $1{,}065$ such pairs, its CF-pair AUROC is $0.498$ (chance $=0.5$).}
\label{fig:teaser}
\end{figure}

Computing grasp poses (grasps) is a fundamental problem in robotic manipulation. Given a grasp on
an object, the end-effector can reach it only if a collision-free path exists from the robot's
start configuration; motion planning is \emph{feasible} when such a path exists and
\emph{infeasible} otherwise. Sampling-based planners~\cite{karaman2011IJRR} commonly used in
manipulation are only probabilistically complete, so they cannot certify infeasibility: they keep
searching until termination or timeout, inflating overall computation time.

Learned predictors~\cite{danielczuk2021scenecollision, yang2023piginet, xu2022nfc, bouhsain2024agfp,
bouhsain2025grn} that score this \emph{feasibility} from a scene are now a standard
speed-up for Task and Motion Planning (TAMP)~\cite{kaelbling2013IJRR, dantam2018IJRR,
garrett2018IJRR, thomas2021RAS}, and on point-cloud benchmarks they report near-perfect
accuracy.\footnote{TAMP combines discrete actions of task planning with the continuous geometric
motion planning. A task-level action may turn out to be infeasible at the motion-level, e.g.,
objects obstructing a grasp, requiring task-level re-planning.} However, these benchmarks evaluate predictor models on naturally sampled grasps, for which the grasp pose is often sufficient to predict the label. A predictor can therefore saturate the benchmark without using the full scene geometry checked by the motion planner. \cref{fig:teaser} shows one such grasp: the initial feasible grasp (left) is made infeasible by inserting the cube (right), yet the predictor's feasibility score remains unchanged. 

Collisions are a frequent and significant source of grasp infeasibility (\cref{sec:setup}). A natural question is \textit{why standard benchmarks fail to reveal collision.} Their sampling rarely includes examples near the boundary where a small change in geometry causes a grasp to switch from feasible to infeasible (\cref{sec:confound}). Consequently, high accuracy on the benchmark does not verify that the model reasons about collisions. The model can achieve high accuracy without actually learning this aspect of grasp feasibility.
This exposes two related issues:
shortcut learning~\cite{geirhos2020shortcut} and causal confusion~\cite{dehaan2019causal}, both of which can be measured directly through controlled interventions.

We evaluate different models using their released weights on their original benchmarks and robots. The models do not fail in the same way. Instead, each responds only at the level of detail its supervision, query formulation, and representation support, and stops below it (\cref{sec:hometurf}, \cref{tab:spectrum}).
 Importantly, the obstacle is present in the input representations. Thus, this insensitivity cannot simply be attributed to PointNet's critical-point-set theorem~\cite{qi2017pointnet}, which concerns the learned per-point representation rather than the geometric structure of the scene. We verify this directly in \cref{sec:why}. Further evidence comes from a clearance-supervised model trained on the same unmodified benchmark, with no counterfactual data at all, which still recovers the collision term (\cref{sec:geometry}). 

\emph{Together, these results show that the relevant collision signal is present in the input, but standard supervision does not require the model to use it.}

\input{sec/1_contribution}

%% file: sec/1_contribution.tex
\noindent\textbf{Contributions.}
We audit eight predictors: three classifiers of a grasp-feasibility
benchmark~\cite{sajid2026feasibility} built on GraspNet-1B scenes~\cite{fang2020graspnet}, and five prior methods spanning
point clouds, voxels, depth images, and object
tokens~\cite{danielczuk2021scenecollision, yang2023piginet, xu2022nfc, bouhsain2024agfp,
bouhsain2025grn}. For~\cite{sajid2026feasibility,bouhsain2025grn,danielczuk2021scenecollision,yang2023piginet}, we additionally evaluate the authors' released weights on their original data and robot platforms. \emph{These results leave open a fundamental question: do the predictors actually capture collision-relevant scene geometry?} We introduce a systematic test, along with metrics and data, to measure this correspondence.

\begin{itemize}[leftmargin=1.2em, itemsep=2pt, topsep=3pt, parsep=0pt]
  \item \textbf{Benchmark saturation and collision limits.}
        Predictors that are near-perfect on the natural benchmark score $0.47$-$0.504$ once we
        insert an obstacle that flips the collision label at the same grasp
        (\cref{sec:saturated}, \cref{tab:saturation}). Natural data almost never places two grasps at the same pose on opposite sides of a contact
        boundary, so a predictor can score well without reading the scene (\cref{sec:confound}).

  \item \textbf{Planner-verified counterfactual test and metrics.}
        We define the \emph{corridor insertion} edit and verify both of its preconditions: every edit
        flips the planner's label, and the inserted obstacle survives the model's downsampling
        (\cref{sec:setup}). We report CF-pair and within-grasp AUROC, and release pick-clear accuracy
        for auditing new predictors. The planner-swap and real-object controls show the flip is not an
        artifact of one planner, arm, or obstacle type (\cref{sec:robust}).

    \item \textbf{Predictors fall short in different ways.}
        Each responds only down to the resolution its representation and supervision support, from a
        bounding box to a $12.5$\,cm voxel, with two point-cloud predictors not resolving the obstacle
        at all (\cref{sec:hometurf}, \cref{tab:spectrum}). Three do so on their authors' own data and
        robot. We therefore report each at its own resolution rather than as one aggregate score.

  \item \textbf{Collision signal in the input and supervised read.}
        A clearance-supervised read of the same points reaches CF-pair $0.983$, trained on the
        unmodified benchmark and never shown a counterfactual scene (\cref{sec:geometry}), consistent
        with ClearanceNet~\cite{kew2020clearancenet}.

  \item \textbf{A released counterfactual benchmark.}
        Each label costs a planner call, so we release the GraspNet-1B
        counterfactual set with planner-verified labels, a datasheet~\cite{gebru2021datasheets}, and
        the relabeling pipeline,\footnote{Repository (code, data, and datasheet):
        \url{https://github.com/codewithsajid/graspfc-cf}.}
        with each obstacle family as its own split. A new predictor can be scored on it without
        running a planner (supp.\ \cref{supp:usage}).
\end{itemize}

%% file: sec/2_related.tex
\section{Related Work}
\label{sec:related}

\noindent \textbf{Learned grasp and motion feasibility.}
Predicting whether a grasp is reachable and collision-free is widely used to prune
candidates before planning, where collision checking is the dominant runtime
cost~\cite{yu2021gnncollision}. GraspNet-1B~\cite{fang2020graspnet} provides real cluttered scenes and
grasp annotations, atop a mature grasp-synthesis
literature~\cite{newbury2023grasp,sundermeyer2021contactgrasp}; a recent benchmark labels its grasps
for motion feasibility with a planner~\cite{sajid2026feasibility}. Feasibility and collision predictors
span point clouds, voxels, depth, and object
tokens~\cite{danielczuk2021scenecollision,murali2023cabinet,yang2023piginet,xu2022nfc,bouhsain2024agfp,kew2020clearancenet,bouhsain2025grn}.
SceneCollisionNet~\cite{danielczuk2021scenecollision} learns an implicit collision function for
rearrangement, queried on a voxelized scene. PIGINet~\cite{yang2023piginet} scores the feasibility of a
whole task-level plan from object tokens, and is never queried on a single approach corridor.
NFC~\cite{xu2022nfc} and AGFP-Net~\cite{bouhsain2024agfp} classify feasibility from depth for
multi-robot manipulation planning. GRN~\cite{bouhsain2025grn} reasons over per-object bounding boxes
and has an explicit obstruction head. Each reports high accuracy on its own benchmark, which we do not
dispute. \textit{We ask a different question: does the benchmark test whether a predictor uses the collision geometry the motion planner checks?}

\noindent \textbf{Benchmark auditing, shortcut learning, and counterfactuals.}
Models that reach high accuracy through a spurious correlate rather than the intended signal, lead to
recurring failure modes~\cite{geirhos2020shortcut,geirhos2019texture}; additional inputs
can make it worse, once it latches onto a non-causal cue~\cite{dehaan2019causal}. Uninformed baselines and
stress tests such as HANS~\cite{mccoy2019hans} expose these shortcuts. Contrast
sets~\cite{gardner2020contrast} and counterfactually-augmented data~\cite{kaushik2020cad} edit input minimally, then re-derive its label. Similar failures recur in 3D vision and robot
learning, e.g., point-cloud networks latching onto rotation-frame cues~\cite{deng2021vectorneurons}. Generalist
policies generalize poorly when the bottleneck is dataset diversity rather than
scale~\cite{xing2025shortcut,emukpere2026brittle}. Our test shows that an uninformed baseline model \emph{can} succeed without the input and therefore may lead to \emph{wrong} predictions when the input alone is changed; a motion planner, not a human annotator, certifies the
new label.

\noindent \textbf{Clearance as a supervised quantity.}
ClearanceNet~\cite{kew2020clearancenet} learns the clearance between a robot configuration and the
scene, and uses the learned estimate to accelerate batched motion planning. Its lesson is that collision geometry is learnable once the query and the label are both stated in terms of clearance. A feasibility
benchmark does not impose that condition. Its binary label supervises the outcome, not the geometric margin that produces it. We show the other half of that picture. Nothing in the benchmark forces a predictor to use the collision term, because a pose shortcut fits its labels just as well. The labels are still sufficient: supervising clearance over the same points, on the benchmark's own splits, recovers the term (\cref{sec:geometry}).

\noindent \textbf{3D encoders and tokenizers.}
Point-cloud recognition is dominated by permutation-invariant encoders with max
pooling~\cite{qi2017pointnet} and their hierarchical~\cite{qi2017pointnetpp,qian2022pointnext} and
transformer~\cite{wu2024ptv3} successors. PointNet's critical-point-set
theorem~\cite{qi2017pointnet} is often read as making such encoders insensitive to added points. The
condition is on the learned per-point map rather than on geometry, and it holds only for a single
global max-pool with no local or positional branch. We measure it directly and find that the inserted
points do move the pooled feature (\cref{sec:why}).  Self-supervised pretraining by masked
reconstruction~\cite{yu2022pointbert} or contrastive objectives~\cite{wu2025sonata} learns recognition
features, which need not encode a contact boundary. Reconstruction tokenizers are the one case where
the obstacle is lost outright (\cref{sec:why}).

%% file: sec/3_setup.tex
\section{Counterfactual Methodology}
\label{sec:setup}
The test requires four components: a fixed grasp, an insertable obstacle in the approach corridor, a planner to recompute the label, and a model input incorporating the obstacle points. With these in place, a single scene edit isolates the collision term while leaving all other factors unchanged. We evaluate on a shared benchmark and within each audited method’s native environment and robot (\cref{sec:hometurf}).

\noindent \textbf{The shared benchmark and its oracle.}
The feasibility benchmark of~\cite{sajid2026feasibility} is built on
GraspNet-1B~\cite{fang2020graspnet}. It has $190$ real RGB-D tabletop scenes with official Seen,
Similar, and Novel object splits. Each grasp carries a label from an
RRT-Connect planner~\cite{kuffner2000rrtconnect} oracle for a 7-DOF Franka Panda in
PyBullet~\cite{coumans2021pybullet}. A grasp is \emph{feasible} only when the planner finds a
collision-free path to the pre-grasp pose (10 cm from grasp pose) from its home configuration. This gives $2{,}706{,}160$ per-grasp labels, $9.6\%$ of
them feasible. The label is effectively deterministic: re-running the planner with $8$ seeds on
held-out grasps gives $0/480$ flips, so label noise is at most $0.6\%$ at $95\%$ confidence. The
oracle is not the variable either. Labels do differ across planners and arms, so we vary both: four
planning algorithms and two kinematic chains, the 7-DoF Panda and a 6-DoF xArm6. We call this the
\emph{planner-swap control} (\cref{sec:regimes}, \cref{tab:robust}).

\noindent \textbf{Collision decides most infeasibility.}
Across all labeled grasps, collision accounts for $64.3\%$ of infeasibility: $49.0\%$ from blocked transit and $15.3\%$ from blocked approach. The remaining $35.7\%$ is due to reachability. We assign each infeasible grasp to the first binding constraint in pipeline order, making these collision rates conservative lower bounds.

\begin{figure}[t]
\centering
\includegraphics[width=\linewidth]{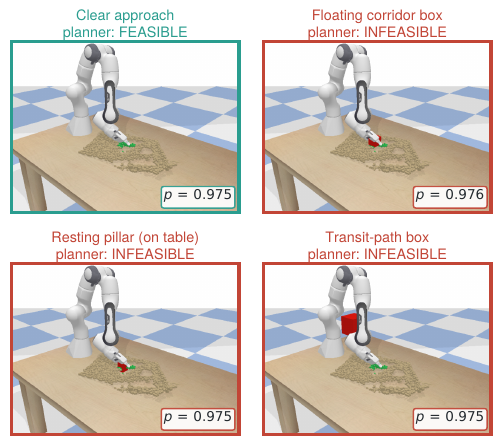}
\caption{The counterfactual protocol on a GraspNet-1B scene (deployed GraspFC-PTX). One grasp is held
fixed; each of three obstacles (a floating corridor box, a resting pillar, a transit-path box) makes the
planner relabel it infeasible (red), while the clear grasp stays feasible (green). The model returns the same
$p(\text{feasible})$, $0.975$ to $0.976$, for all four, so its score does not separate the clear scene
from any of the three blocked ones.}
\label{fig:gallery}
\end{figure}

\noindent \textbf{The edit.}
A perturbation is useful only if it changes the label and leaves a physically valid scene. We hold the
grasp pose fixed, drop a solid box into the $10$\,cm approach corridor, re-run the planner for the true
new label, and add the box points to the model input (\cref{fig:gallery}). We call this edit a
\emph{corridor insertion}. The box has edge $1.5$ to $5$\,cm and sits at a $5$ to $15$\,cm standoff.
Two properties are verified. First, the edit
causes a real flip: it relabels every pair, while a matched box placed $30$\,cm off-corridor leaves the
grasp feasible $98.0\%$ of the time. Second, the obstacle survives downsampling, since every item
retains at least one box point, $\approx 13$ on average, after farthest-point sampling to the
$2048$-point input. A graded variant slides the box through six corridor offsets, from blocking to
harmless, so a model must track \emph{where} the obstacle is.

\noindent \textbf{Obstacle families.}
The corridor box is the primary family, but we also vary the geometry, including spheres, non-convex L-shapes, table-mounted pillars, transit-path boxes, and four obstacle clusters. Size, standoff, and placement are independently swept, spanning six corridor offsets from blocking to harmless, plus adjacent, distal, and on-axis placements. A no-primitive family, the \emph{real-object control}, uses a real benchmark object placed beside the target. Each family is planner-verified and released as a separate audit split (\cref{sec:robust}, \cref{sec:ood}, supp.\ \cref{supp:usage}); geometry
and the sampling details are specified in supp.\ \cref{supp:datasheet}.

\noindent \textbf{Metrics.}
We report \emph{natural AUROC} on the original splits without inserted obstacles. \emph{CF-pair AUROC} ranks clear versus blocked versions of the same grasp at an identical pose, while \emph{within-grasp AUROC} ranks feasible versus infeasible obstacle offsets. Chance is $0.5$ for both. The audit suite additionally reports \emph{pick-clear} accuracy: among two candidates where one is blocked, how often the clear grasp is ranked first (supp.\ \cref{supp:usage}). Pose-only models cannot exceed chance on any metric, since paired examples share the same pose. CF-pair AUROC uses $1{,}065$ released pairs, with bootstrap $95\%$ CIs over pairs; trained results use $3$ seeds. Thus, scores near $0.5$ reflect genuine failure to detect collision, not estimation noise.

\noindent \textbf{Models and evidence regimes.}
We evaluate three model groups (\cref{tab:saturation}). \textbf{(i)} The three GraspFC classifiers
of~\cite{sajid2026feasibility}, run from their released weights on the shared benchmark. These are
GraspFC-NNet, -Conv3D, and -PTX, the last on a PTv3~\cite{wu2024ptv3} backbone. \textbf{(ii)} Four
prior representations, re-implemented and trained on the planner labels under one protocol:
SceneCollisionNet, PIGINet, NFC~\cite{xu2022nfc}, and AGFP-Net~\cite{bouhsain2024agfp}. They share the
oracle, the splits, and the loss, so this group tests the representation rather than the authors'
deployed system. \textbf{(iii)} Released weights of published methods run on their own data
(\cref{sec:hometurf}). Per-model architectures and losses are in supp.\ \cref{supp:training}. Every label on
the shared benchmark comes from the same RRT-Connect oracle, which we verify deterministic and causal
(supp.\ \cref{supp:neg}).

%% file: sec/4_results.tex
\section{Experiments}
\label{sec:results}
\subsection{Benchmark saturation and collision sensitivity}
\label{sec:saturated}

\cref{tab:saturation} reports the audited predictors. All are near-perfect on the natural benchmark, but this score does not distinguish predictors that use collision geometry from those that do not. On same-pose counterfactuals, every predictor scores $0.47$-$0.504$ AUROC (within-grasp $0.47$ to $0.52$). GRN is the one predictor whose score changes at all, and only to obstacle presence: it cannot resolve the sub-bounding-box shape that determines collision
(\cref{fig:hometurf}), so its entry ($^{*}$) is a shape swap rather than an insertion. This failure is not due to input modality. The obstacle appears in the voxel grid of GraspFC-Conv3D and in all five depth views of AGFP-Net, yet both discard it; NFC uses only the grasp pose and never its neighbor channel
(\cref{tab:spectrum}). The deciding geometry is therefore present in the inputs and can be recovered from the same points (\cref{sec:geometry}).

\begin{table}[t]
\centering
\small
\setlength{\tabcolsep}{3pt}
\resizebox{\columnwidth}{!}{%
\begin{tabular}{@{}llrlrr@{}}
\toprule
Predictor & Input & Par.\ & Regime & Nat.\ & CF-pair \\
\midrule
GraspFC-NNet~\cite{sajid2026feasibility} & pose+crop & 1.65M & released & 0.995 & 0.491 \\
GraspFC-Conv3D~\cite{sajid2026feasibility} & pose+voxels & 9.7M & released & 0.976 & 0.497 \\
GraspFC-PTX~\cite{sajid2026feasibility} & pose+cloud & 9.0M & released & \textbf{0.996} & 0.498 \\
SceneCollisionNet~\cite{danielczuk2021scenecollision} & voxels & 26M & re-impl.+rel. & 0.964 & 0.470 \\
PIGINet~\cite{yang2023piginet} & tokens & 31M & re-impl.+rel. & 0.960 & 0.500 \\
NFC$^{\dagger}$~\cite{xu2022nfc} & depth & 22.8k & re-impl. & 0.965 & 0.504 \\
AGFP-Net$^{\dagger}$~\cite{bouhsain2024agfp} & depth views & 33M & re-impl. & 0.920 & 0.498 \\
\midrule
GRN~\cite{bouhsain2025grn} & box+pose & 2.3M & released & 0.995 & 0.500$^{*}$ \\
\bottomrule
\end{tabular}}
\caption{The eight audited predictors: input modality, parameters, evidence regime, and natural vs.\
counterfactual AUROC on the shared benchmark (GRN on its own, below the rule). \emph{released}: the
authors' released weights; \emph{re-impl.}: re-implemented and trained on the planner labels
(``+rel.'': also run from released weights on its authors' own data, \cref{tab:spectrum}).
$^{\dagger}$No public code or weights; re-implemented from the published description.
\emph{Nat.}: ranking feasible vs.\ infeasible grasps on the unmodified benchmark. \emph{CF-pair}:
ranking the clear vs.\ obstacle-inserted version of a grasp at the identical pose, on the
planner-verified pair set ($n{=}1{,}065$); $0.5$ is the no-signal baseline. $^{*}$GRN detects presence but
does not resolve sub-bounding-box shape, so its entry is the same-bounding-box shape swap
(\cref{tab:spectrum}). The methods' original (non-AUROC) headlines and full interventional scores are
in supp.\ \cref{supp:tab:cards,supp:taxonomy}.}
\label{tab:saturation}
\end{table}

\noindent \textbf{Why natural labels do not require collision.}
\label{sec:confound}
Feasibility labels are geometrically separated, with labels flipping at contact boundaries that natural sampling rarely straddles, making same-pose contrast essentially absent. Conditioning on grasp difficulty, which carries no scene information, therefore removes most of the model's apparent advantage. This matches the benchmark's own observation
that grasp pose rather than scene geometry carries the feasibility
signal~\cite{sajid2026feasibility}. The \emph{input} holds the collision geometry, as shown in
\cref{sec:geometry}, recovering it
under the benchmark's own labels. The \emph{label distribution} never requires a predictor to use it,
so no quantity of natural data forces the term into the model. This follows from how the labels distribute, not from one benchmark's design.

\subsection{Robustness to placement, obstacle, planner, and robot}
\label{sec:regimes}
\begin{figure}[t]
\centering
\includegraphics[width=\linewidth]{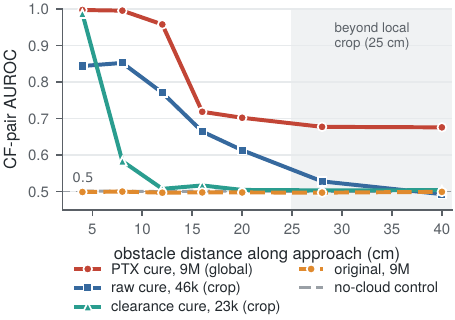}
\caption{CF-pair AUROC vs.\ obstacle distance along the approach (inference-only).
The three counterfactual-cured variants (the deployed global PTX, and two crop-input models using raw
corridor points vs.\ clearance-aware tokens) react only near the gripper; the original model and a
no-cloud control stay at $0.5$.}
\label{fig:cliff}
\end{figure}

The result is not an artifact of our re-implementations or of one obstacle location. Re-scoring \emph{inference-only}, with no model seeing a new label, we move the obstacle to held-out
scenes and unseen placements. These are an adjacent in-crop offset (CF-adjacent, $n=561$) and a distal
out-of-crop offset (CF-distal, $n=250$). Every model stays in $[0.47, 0.554]$ at every placement, and
its score barely moves with or without the obstacle (\cref{tab:placement}). \cref{fig:cliff} shows the
dual fact for the \emph{cured} models, those we fine-tune or train on planner-relabeled counterfactual
pairs. Counterfactual training makes them react only within a shell around the gripper. No model we
test, cured or not, holds a clearance representation that survives moving the obstacle across the
workspace. The result also holds across standoff and obstacle type, shown below, and across
dense multi-object clutter (supp.\ \cref{supp:fig:clutter}). Most tellingly, it holds in the real-object control, where the obstacle
is not a foreign primitive at all: a GraspNet-1B object from the same Novel split, resting
on the table next to the target, flips every one of $150$ grasps, while a matched copy set aside leaves
all of them feasible. The deployed model returns essentially the same score either way
($|\Delta p| = 0.001$; \cref{tab:robust}, supp.\ \cref{fig:indist}). The diagnosis is therefore not an
out-of-distribution obstacle artifact.

\begin{table}[t]
\centering
\small
\setlength{\tabcolsep}{4pt}
\begin{tabular}{@{}lrrrr@{}}
\toprule
Model (run as-is) & CF-iid & CF-adj.\ & CF-dist.\ & max $|\Delta p|$ \\
\midrule
GraspFC-PTX (9.0M) & 0.501 & 0.499 & 0.497 & $<$0.001 \\
PIGINet & 0.500 & 0.508 & 0.493 & 0.014 \\
NFC & 0.504 & 0.511 & 0.532 & 0.014 \\
AGFP-Net & 0.498 & 0.497 & 0.499 & 0.007 \\
SceneCollisionNet (26M) & 0.470 & 0.473 & 0.554 & 0.052 \\
\bottomrule
\end{tabular}
\caption{Placement-invariance (inference-only, no retraining). CF-pair AUROC with the obstacle at the
in-distribution corridor offset (iid), an unseen adjacent offset ($n{=}561$), and a distal offset
outside the crop ($n{=}250$), all in held-out scenes. The one cell above $0.5$ at distance
(SceneCollisionNet, $0.554$) is $\approx 1.4$\,SE from $0.5$ and falls below it at the other two
offsets. max $|\Delta p|$ is the largest score change the obstacle produces.}
\label{tab:placement}
\end{table}

\noindent \textbf{Oracle, obstacle type, and embodiment.}
\label{sec:robust}
We re-run the planner-verified counterfactual under three standoffs and two further obstacle types,
and with an in-distribution dataset object from the test splits placed in the grasp approach direction
(supp.\ \cref{fig:indist}). The deployed model is re-scored on each (\cref{tab:robust}). The
\emph{resting} case is gravity-plausible: a pillar standing on the table that rises into the descent
corridor, not a floating body. The third case lies in the arm's \emph{transit} path, the dominant
$49.0\%$ blocked-transit failure mode. The edits therefore span both feasibility terms and are
physically realizable. In every variant the planner relabels the grasp, with causal flip $0.56$ to
$1.00$, while a matched off-corridor null box leaves it feasible $\ge 97\%$ of the time. The deployed
model stays near chance throughout (CF-pair AUROC $0.491$ to $0.500$) and returns an essentially
identical score with and without the obstacle ($|\Delta p| \le 0.001$). Dense clutter behaves the same:
a cluster of four obstacles flips all $150$ re-checked grasps, and the model stays at chance (CF-pair
$0.498$, $|\Delta p| < 0.001$; supp.\ \cref{supp:fig:clutter}).

\begin{table}[t]
\centering
\small
\setlength{\tabcolsep}{4pt}
\begin{tabular}{@{}lrrrr@{}}
\toprule
Oracle / obstacle & Flip & Null & CF-pair & $|\Delta p|$ \\
\midrule
Corridor box, $10$\,cm standoff & 1.00 & 0.02 & 0.498 & 0.001 \\
Corridor box, $5$\,cm standoff & 1.00 & 0.02 & 0.499 & 0.000 \\
Corridor box, $15$\,cm standoff & 1.00 & 0.02 & 0.500 & 0.000 \\
Resting pillar (on the table) & 0.74 & 0.03 & 0.499 & 0.000 \\
Transit-path box & 0.56 & 0.01 & 0.491 & 0.001 \\
Dataset object (on the table) & 1.00 & 0.00 & 0.500 & 0.001 \\
\midrule
\multicolumn{5}{@{}l@{}}{\emph{Planner swap} ($4$ planners $\times$ $2$ arms, $8{,}412$ planner calls):} \\
\multicolumn{5}{@{}l@{}}{flip $1.000$ in $8$ of $8$ conditions; label agreement $0.973$--$0.995$;} \\
\multicolumn{5}{@{}l@{}}{Cohen's $\kappa$ $0.940$--$0.990$.} \\
\bottomrule
\end{tabular}
\caption{Robustness of the flip and of the model's insensitivity to oracle standoff, obstacle type,
planner, and arm. \emph{Flip}: fraction of grasps the inserted obstacle relabels infeasible (planner
ground truth). \emph{Null}: false-flip rate of a matched off-corridor box that should stay feasible
(ideal $0$). \emph{CF-pair} (chance $0.5$) and $|\Delta p|$ (score change with vs.\ without the
obstacle) are for the deployed model (GraspFC-PTX). The planner relabels every variant while the model
stays at chance. Labels themselves differ across planners and arms ($607$ of $957$ Panda-feasible
grasps are out of xArm6 reach); the flip does not. Per-condition values: supp.\ \cref{supp:tab:planners}.}
\label{tab:robust}
\end{table}

The oracle itself is not the variable either. The planner-swap control relabels the grasp in $8$ of
$8$ conditions at flip rate $1.000$ (\cref{tab:robust}). The diagnosis therefore tracks the
representation, not the standoff, the obstacle shape, the planning algorithm, or the arm.
\subsection{Released weights on the authors' own data}
\label{sec:hometurf}

\begin{figure}[t]
\centering
\includegraphics[width=\linewidth]{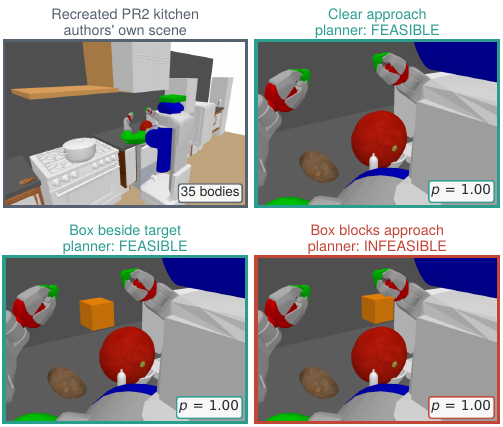}
\caption{PIGINet on the authors' own PR2 kitchen ($35$-body saved scene, top-left; released weights). At
one fixed arm pose: a clear approach, a harmless box beside the target, and a box blocking the approach.
The planner relabels only the blocking case infeasible (red), yet PIGINet returns the identical $p=1.00$
in all three, so its score does not register the corridor box.}
\label{fig:kitchen}
\end{figure}

\begin{figure}[t]
\centering
\includegraphics[width=\linewidth]{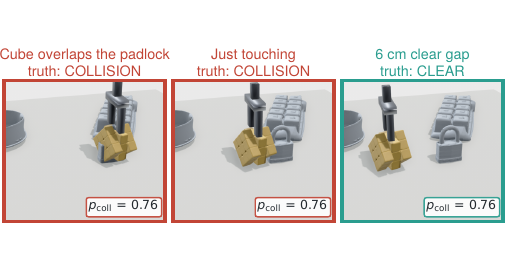}
\caption{SceneCollisionNet on a cluttered ACRONYM-style tabletop, released checkpoint. A gripper-held
cube is swept against a solid obstacle from overlapping it, through contact, to a $6$\,cm clear gap. The
true label flips (red to teal), yet the score holds at $p_{\mathrm{coll}}=0.76$ ($0.75$ to $0.77$ over
the sweep, never below $0.5$), where an ideal checker steps from $1$ to $0$. The whole sweep is shorter
than the $12.5$\,cm voxel it reasons over.}
\label{fig:scn}
\end{figure}

\begin{figure*}[t]
\centering
\includegraphics[width=\linewidth]{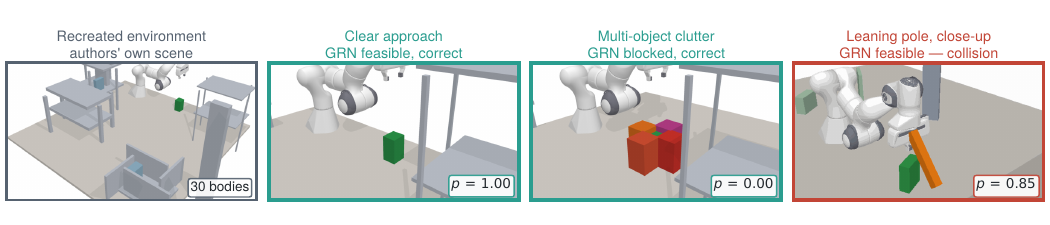}
\caption{GRN on its own panda\_3d\_4 scenes ($30$-body environment, left; released weights). GRN
detects inserted geometry, dropping from feasible $1.00$ (clear) to $0.00$ when a multi-object cluster
blocks the approach. What it does not resolve is sub-bounding-box shape (rightmost, close-up from a
rollout): a thin pole leaning $40^\circ$ into the approach corridor is read as an upright box and called
feasible ($p\,{=}\,0.85$), yet the gripper drives straight through it. GRN reads the bounding box,
not the shape (\cref{tab:spectrum}).}
\label{fig:hometurf}
\end{figure*}

Most directly, we run published models from their \emph{own released weights on their own benchmarks}
(\cref{tab:spectrum,fig:kitchen,fig:scn,fig:hometurf}), which removes the re-implementation caveat. The
verdicts are not uniform. Each model responds at the resolution its supervision, native query
semantics, and representation provide, and stops responding below it (\cref{tab:spectrum}). PIGINet, on its authors' PR2 kitchen, calls a planner-blocked pick feasible $90\%$ of the time against
a ground truth of $0\%$ (\cref{fig:kitchen}; aggregated in supp.\ \cref{supp:fig:kitchen_agg}). It is a
sequence-level predictor over object tokens and was never queried on an approach corridor, and the
shortcut extends to multi-step plans (supp.\ \cref{supp:seq}). SceneCollisionNet, on its own
ACRONYM~\cite{eppner2021acronym} scenes, is the one audited predictor that responds to shape above
chance ($0.821$ on a same-bounding-box swap). It is not a decision boundary. Swept from a deep overlap
to a $6$\,cm clear gap, its score holds at $p_{\mathrm{coll}}=0.76$ straight across the true contact
boundary (\cref{fig:scn}). It reads contact, and stops at its $12.5$\,cm voxel (supp.\ \cref{supp:scn}). GRN reads a bounding box by design.
It \emph{detects} inserted geometry, yet scores a box and its inscribed same-bounding-box ellipsoid
identically: a \emph{shape-swap AUROC} of $0.500$, with two further geometry holes in supp.\ \cref{supp:grn}.
One aggregate number over these predictors would hide five mechanisms with five different remedies.

\begin{table}[t]
\centering
\small
\setlength{\tabcolsep}{2pt}
\scalebox{0.78}{
\begin{tabular}{@{}lllrr@{}}
\toprule
Predictor & Responds to & Stops below & Native & Interv.\ \\
\midrule
GRN$^{\dagger}$ & presence, bbox & sub-bbox shape & 0.995 & 0.500 \\
SceneCollisionNet$^{\dagger}$ & contact, coarsely & $12.5$\,cm voxel & 0.969 & \textbf{0.821} \\
PIGINet$^{\dagger}$ & object tokens & corridor geometry & 0.86 & 0.470 \\
\midrule
GraspFC-NNet & presence & whether it blocks & --- & 0.491 \\
GraspFC-Conv3D & voxel grid & any obstacle & --- & 0.497 \\
GraspFC-PTX & (no response) & any obstacle & --- & 0.498 \\
NFC & grasp pose only & its neighbor channel & --- & 0.504 \\
AGFP-Net & $5$-view depth & any obstacle & --- & 0.498 \\
\midrule
Clearance tokens (ours) & corridor clearance & off-family placement & --- & \textbf{0.983} \\
\bottomrule
\end{tabular}}
\caption{The physical-sensitivity spectrum on one protocol, for all eight audited predictors.
\emph{Responds to} is the coarsest geometry that moves the score; \emph{stops below} is where the
response no longer tracks the planner. $^{\dagger}$Released weights on the method's own data and
robot: \emph{Native} is our AUROC, \emph{Interv.}\ the counterfactual score on that same set (GRN
and SceneCollisionNet: shape swap; PIGINet: insert-a-box). Each reproduces its authors' published
headline first, none of which is an AUROC: GRN's action, grasp, and IK F1 and obstruction
MAE~\cite{bouhsain2025grn}, SceneCollisionNet's trajectory accuracy and
AP~\cite{danielczuk2021scenecollision}, PIGINet's validation plan-feasibility and planning-time
cut~\cite{yang2023piginet}; values in supp.\ \cref{supp:tab:cards}. The remaining rows are on the
shared benchmark, where \emph{Native} is \cref{tab:saturation}'s Nat.\ column, so \emph{Interv.}\
compares within a regime and not across. GraspFC-NNet and GraspFC-PTX differ by $28\times$ in how
strongly the obstacle moves their score ($|\Delta p|$ = $0.027$ / $0.001$), yet neither tracks collision. Per-model mechanisms are in supp.\ \cref{supp:tab:taxonomy}.}
\label{tab:spectrum}
\end{table}

\subsection{Where the collision term is lost}
\label{sec:why}

If the deciding evidence is in the input, where is it lost? Not at the pooling step, and not by
construction. PointNet's critical-point-set theorem~\cite{qi2017pointnet} does not say that points
inside a critical set's hull cannot change the global feature. It gives a critical set and an
upper-bound set, and adding points leaves the pooled feature unchanged exactly when no added point
raises a pooled channel. That is a condition on the learned per-point map, not on geometry. We measure
it on GraspFC-NNet's max-pooled branch in isolation, over the planner-verified pairs, and the condition
fails. Only $16\%$ of the inserted points raise no channel, and the obstacle raises $31.5$ of $256$
channels. The pooled feature moves \emph{more} than a mean pool would move it ($0.174$ against
$0.099$), because the inserted points are extremal.

The obstacle therefore enters the representation. Decomposing the deployed forward pass shows how far
it gets. Inside the full model that same max-pooled branch moves by $0.074$ and the scene branch by
$0.212$. The head is not frozen either, with a per-pair $|\Delta\text{logit}|$ of $0.365$. What the
movement lacks is sign. The mean $\Delta\text{logit}$ is only $+0.017$, and the alignment between the
feature displacement and the one direction the head can read has mean cosine $+0.043$ against a
magnitude of $0.251$. The response is large and direction-random. We test this directly by placing two
obstacles at the identical grasp pose, adding the same number of points to the same cloud, one blocking
the approach and one $30$\,cm off-corridor. GraspFC-NNet registers both strongly ($|\Delta p|$ $0.027$)
and GraspFC-PTX registers neither ($0.001$). Neither ranks the blocking one below the harmless one
(blocking AUROC $0.481$ and $0.502$, bootstrap intervals containing $0.5$). These are two different
failures with two different fixes.

Nor is the cause the weights (\cref{tab:encoders}). Neither recognition-pretrained encoders
(PointNet++~\cite{qi2017pointnetpp}, PointNeXt-S~\cite{qian2022pointnext}) nor the deployed model's own
feasibility-trained PTv3~\cite{wu2024ptv3} lets clearance be linearly read from its features. Only an
explicit clearance tokenization of the same points does. Pretraining is irrelevant: GraspFC-PTX is
near-perfect on the natural benchmark yet at $0.498$ CF-pair, whether trained from scratch or from
Sonata self-supervised weights~\cite{wu2025sonata}. Reconstruction does not help either. The deciding
obstacle is a small fraction of the points, so a Point-BERT-style
autoencoder~\cite{yu2022pointbert} discards most of it on encoding. Nothing in the natural supervision
ever required the score to move in the right direction, which is the sense in which the term is not
extracted. Clearance supervision changes that (\cref{sec:geometry}).

\begin{table}[t]
\centering
\small
\setlength{\tabcolsep}{5pt}
\begin{tabular}{@{}llrr@{}}
\toprule
Encoder & Supervision & $\Delta$feat.\ & Probe $R^2$ \\
\midrule
PointNet++ & recognition & $-0.014$ & 0.03 \\
PointNeXt-S & recognition & $+0.001$ & 0.05 \\
PTv3 & feasibility & $+0.002$ & 0.01 \\
Clearance tokens (ours) & clearance & $\mathbf{+1.48}$ & \textbf{0.71} \\
\bottomrule
\end{tabular}
\caption{Frozen-encoder response to a feasibility-flipping obstacle. \emph{Supervision} is what each
encoder was trained to encode (recognition on ModelNet40~\cite{wu2015modelnet}, feasibility, or
clearance); PTv3 is the deployed model's own backbone, so it is the in-domain control. $\Delta$feat.\ is
the pooled-feature change in native units (within-row only, \emph{not} comparable across rows); Probe
$R^2$ is the cross-encoder readout of how well a linear map recovers corridor clearance. Only the
clearance tokenizer both responds and is readable. These are frozen encoders probed in isolation, and
are not comparable to the deployed forward-pass displacements quoted in the text.}
\label{tab:encoders}
\end{table}

\subsection{What does not close the gap, and what does}
\label{sec:fixes}

Nothing on the natural benchmark closes the gap (\cref{tab:fixes}). Scaling natural data does nothing.
Fine-tuning the deployed model on counterfactuals does not transfer across five standard recipes, and
width is not the lever either, since a $10\times$ wider encoder tracks the baseline. The shortcut is
imprinted in the pretrained weights. The obvious escape is to retrain the same architecture from
scratch on counterfactual labels. That does raise the in-distribution counterfactual score (CF-pair
$0.991$), which alone shows the limit is not representational in principle. But it is \emph{fitting
the intervention} rather than learning clearance. The cured model is placement-bound, and its score
falls back toward chance the moment the obstacle leaves the trained corridor (\cref{fig:cliff}), so
retraining on a new placement only fits a new shortcut. Counterfactual data buys coverage of the
trained obstacle distribution, not a transferable collision representation. The lever is the
representation, which we turn to next.

\begin{table}[t]
\centering
\small
\setlength{\tabcolsep}{3pt}
\scalebox{0.9}{
\begin{tabular}{@{}lrll@{}}
\toprule
Intervention & Nat.\ & CF-pair & Verdict \\
\midrule
More natural data & 0.990 & 0.498 & term not identified \\
Fine-tune deployed & 0.990 & 0.495 to 0.508 & does not transfer \\
Train from scratch on CF & 0.990 & \textbf{0.991} & fits intervention \\
\bottomrule
\end{tabular}}
\caption{More data does not teach a transferable collision term; the from-scratch cure fits the
intervention and is placement-bound (\cref{fig:cliff}).}
\label{tab:fixes}
\end{table}

\begin{figure}[t]
\centering
\includegraphics[width=\linewidth]{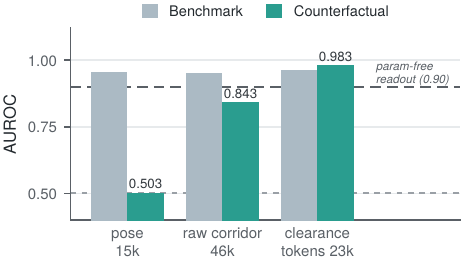}
\caption{Train-time learning on the unmodified benchmark (plain BCE, no counterfactual scenes). Three
representations of the same corridor points, all high on the benchmark; only the clearance-aware
tokenization separates under the protocol (CF-pair $0.983$). Dashed: the param-free geometric readout.}
\label{fig:geom}
\end{figure}

\noindent \textbf{Clearance supervision on the unmodified benchmark.}
\label{sec:geometry}
Concretely, the collision term can be learned at train time from the unmodified benchmark alone, with
no counterfactual scenes, by exposing its geometric signal (corridor clearance) to the network.
\cref{fig:geom} shows the ladder, all trained on the original splits with plain BCE. A pose
representation stays at chance (CF-pair $0.503$). Raw corridor points recover part of the term
($0.843$). A clearance-aware tokenization of the \emph{same} points reaches CF-pair $0.983$ with only
$23$k parameters, beating a param-free geometric readout of those tokens ($0.90$). The tokenization
slices the $10$\,cm corridor into $K{=}13$ axial slabs (supp.\ \cref{fig:ict}). This model is trained
on the unmodified benchmark, and is distinct from the from-scratch $0.991$ of \cref{tab:fixes}, which
trains on \emph{counterfactual} data. Two further train-time changes sharpen it on the same data: a
per-point clearance channel, and a soft-min over per-region clearances that makes the predictor
monotone by construction (supp.\ \cref{supp:ict}). Two models tied on the benchmark thus separate under the
protocol on no new data. The separation comes from a \emph{learned} read of the clearance tokens, not
an analytic check bolted onto the output. This is a diagnostic, not a deployable predictor. Like the
from-scratch cure it still
degrades off the trained placement distribution, as we show next, so a clearance representation that
\emph{generalizes} across placement (supp.\ \cref{supp:ict}) is future work.

\begin{figure}[t]
\centering
\includegraphics[width=\linewidth]{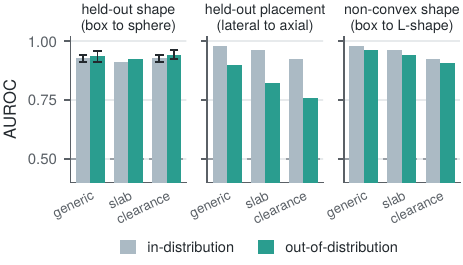}
\caption{Out-of-distribution transfer of the three cured models of \cref{sec:geometry} (generic, slab,
clearance). Each panel is a held-out obstacle family, planner-relabeled; bars are AUROC in- versus
out-of-distribution and the dashed line is chance ($0.5$). Cures transfer at or near their in-distribution
score; box-to-sphere error bars are bootstrap $95\%$ CIs ($n{=}532$ ID, $250$ OOD).}
\label{fig:ood}
\end{figure}

\noindent \textbf{Transfer to unseen obstacle shapes and placements.}
\label{sec:ood}
The hostile reading is that a counterfactual-cured model merely memorized our obstacle generator. It
did not: trained on box counterfactuals and evaluated, planner-relabeled, on never-seen \emph{spheres},
the cure transfers at or above its in-distribution score, while a no-cloud control stays at chance. It
also holds on a never-seen \emph{non-convex} L-shape and transfers, more weakly, to an unseen on-axis
\emph{placement} family, in every case far above chance. The cure computes occupancy from the points, so it transfers across held-out obstacle shape and, more
weakly, placement, rather than memorizing one generator. Per-architecture scores and bootstrap CIs are
in \cref{fig:ood}.

%% file: sec/5_conclusion.tex
\section{Discussion and Conclusion}
\label{sec:conclusion}

We audit grasp-feasibility predictors with the \emph{corridor insertion}: a planner-verified counterfactual edit
that holds the grasp pose fixed, places a solid obstacle in the approach corridor, and enters its
points into the model input. It flips the label on every pair and survives the model's
downsampling. Collision decides $64.3\%$ of
infeasibility, but predictors with near-perfect natural AUROC score $0.47$ to $0.504$ on the
counterfactual pairs. The obstacle does reach the representation: in GraspFC-NNet its scene
feature moves by $21\%$. The score still does not change with whether the obstacle blocks. The
natural
distribution almost never contains the fixed-pose contrast that would require it to
(\cref{sec:confound}). The input carries the collision term, and the benchmark does not
identify it.

The failure is not uniform. GRN reads a bounding box, SceneCollisionNet a $12.5$\,cm voxel,
GraspFC-NNet registers the obstacle but cannot say whether it blocks, and GraspFC-PTX, the
deployed model, does not register it at all ($|\Delta p| \approx 0.001$). Each fails below its
own resolution. This holds across placement, dense clutter, the real-object control, and every
planner-swap condition. Clearance supervision closes the gap: the benchmark's own labels
are sufficient, and a $23$k-parameter read of corridor clearance reaches CF-pair $0.983$
(\cref{sec:geometry}), while training on the interventions instead only fits them. That read stays
bound to its trained placement distribution, and we evaluate in simulation with
a sampling-based planner as oracle. We release the pairs, metrics, datasheet, and pipeline, so
new predictors can be audited without a planner call (supp.\ \cref{supp:usage}). Generalizing a
clearance representation across obstacle placement remains an open problem.

%% file: sec/6_supp.tex
\noindent\textbf{Roadmap.} We first give the robustness figures (oracle standoff,
obstacle type, a dataset object, and dense clutter) and the physical cost of the result;
\cref{supp:seq} then extends the protocol to multi-step plan feasibility; \cref{supp:scn} gives the
shape- and contact-resolution detail for GRN and SceneCollisionNet on their own benchmarks; \cref{supp:training}
lists per-model architecture, training pool, and in-distribution accuracy; \cref{supp:taxonomy} gives
the per-model interventional scores and the representation-to-failure taxonomy; \cref{supp:grn}
dissects why GRN partially passes; \cref{supp:neg} reports the negative controls and label-integrity
checks; \cref{supp:ict} specifies the clearance-aware tokenization; and
\cref{supp:datasheet,supp:usage} provide the datasheet and the audit protocol for the released benchmark.

\setcounter{section}{4}
\section*{Robustness of the diagnosis: figures and the physical cost}
\label{supp:robustsec}
\noindent The main paper establishes that the diagnosis survives the oracle standoff, the obstacle type, the
planner, and the arm (\cref{tab:robust}). Here we give the figures behind that result, the
per-condition planner-swap scores, a worked example, dense clutter, and the downstream cost of the
result.

\begin{figure}[t]
\centering
\includegraphics[width=\linewidth]{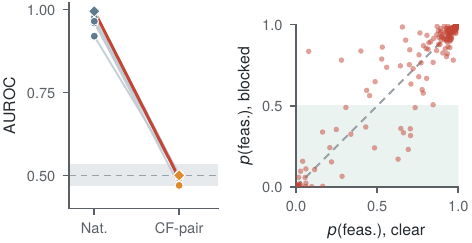}
\caption{\emph{Left}: every predictor in \cref{tab:saturation} falls from
near-perfect natural AUROC to near $0.5$ under the same-pose counterfactual (one line each;
GraspFC-NNet red). \emph{Right}: for GraspFC-NNet, the blocked grasp gets essentially the clear grasp's
score, so the same-pose pairs whose label flips lie on the $y{=}x$ diagonal rather than in the region a
collision-aware model would occupy (shaded).}
\label{supp:fig:overlap}
\end{figure}

\begin{figure}[t]
\centering
\includegraphics[width=\linewidth]{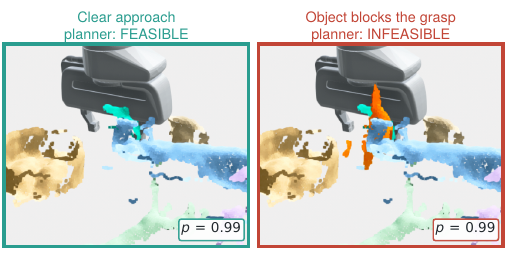}
\caption{The obstacle is a GraspNet-1B object (from the scene segmentation, same Novel split)
resting on the table next to the target. The planner relabels the grasp infeasible; the deployed SOTA
returns the identical score. Over $150$ such pairs the planner flips every one while the model cannot
tell the two apart (CF-pair AUROC $0.5$): the result extends to dataset objects, not only inserted
boxes.}
\label{fig:indist}
\end{figure}

\begin{table}[t]
\centering
\small
\setlength{\tabcolsep}{4pt}
\begin{tabular}{@{}llrrr@{}}
\toprule
Robot & Planner & Agree & $\kappa$ & Flip \\
\midrule
Panda & BiRRT-Connect, re-run (control) & 0.995 & 0.990 & \textbf{1.000} \\
Panda & BiRRT-Connect, other hyper-params & 0.990 & 0.977 & \textbf{1.000} \\
Panda & single-tree goal-biased RRT & 0.973 & 0.940 & \textbf{1.000} \\
Panda & Lazy-PRM & 0.992 & 0.983 & \textbf{1.000} \\
\midrule
xArm6 & BiRRT-Connect, re-run (control) & 0.988 & 0.973 & \textbf{1.000} \\
xArm6 & BiRRT-Connect, other hyper-params & 0.989 & 0.975 & \textbf{1.000} \\
xArm6 & single-tree goal-biased RRT & 0.984 & 0.964 & \textbf{1.000} \\
xArm6 & Lazy-PRM & 0.989 & 0.975 & \textbf{1.000} \\
\bottomrule
\end{tabular}
\caption{What the causal edit does to the label, under four planning algorithms and two kinematic
chains. The labels themselves are not identical across conditions; the flip is. Only the planner object
or the robot changes; the scene, the grasp poses, the contact tolerance, and the $21$-step approach
discretization are the same code. \emph{Agree} is label agreement with the released oracle and
$\kappa$ is Cohen's $\kappa$, both over the clear, causal, and null conditions. \emph{Flip} is the
fraction of causal edits the planner relabels infeasible. The causal edit flips the label in $8$ of
$8$ conditions without exception, over $8{,}412$ planner calls. Residual disagreement sits on the
clear and null branches and is bounded by the RNG-only control in the first row of each block. The
xArm6 is the harder embodiment: $607$ of $957$ Panda-feasible candidate grasps are out of its reach,
and every corridor insertion still flips the label.}
\label{supp:tab:planners}
\end{table}

\paragraph{A worked example.}
\cref{fig:gallery} in the main paper shows one such instance in full: a single grasp on a
GraspNet-1B scene under all four conditions. \cref{supp:tab:gallery} reports its per-condition scores. The
three obstacles trigger two distinct geometric failure modes (a blocked approach for the corridor box
and the resting pillar, a blocked transit for the path box), and the deployed SOTA returns the same
$p(\text{feasible})$ to within $0.001$ across all four, although the planner flips three of them. This
is not a hand-selected pair: it is one of the instances counted in \cref{tab:robust}, where the
planner flipped every obstacle type while the model never moved.

\begin{table}[t]
\centering
\small
\setlength{\tabcolsep}{5pt}
\begin{tabular}{@{}llll@{}}
\toprule
Condition & Planner & Failure mode & model $p$ \\
\midrule
Clear approach & feasible & none & 0.975 \\
Floating corridor box & infeasible & blocked approach & 0.976 \\
Resting pillar & infeasible & blocked approach & 0.975 \\
Transit-path box & infeasible & blocked transit & 0.975 \\
\bottomrule
\end{tabular}
\caption{Per-condition scores for the worked example of \cref{fig:gallery}: one grasp under four conditions
(deployed SOTA, released weights).}
\label{supp:tab:gallery}
\end{table}

\begin{figure}[t]
\centering
\includegraphics[width=\linewidth]{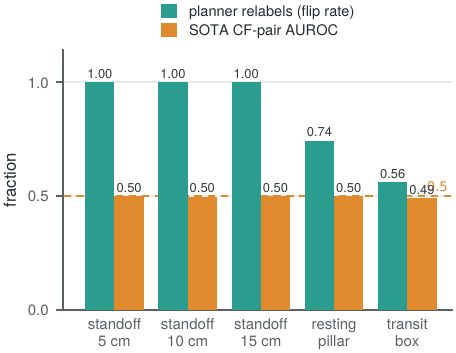}
\caption{\cref{tab:robust} as a chart: planner relabel rate (teal) versus deployed-SOTA CF-pair
AUROC (amber) across three standoffs and two further obstacle types (a gravity-resting pillar and a
transit-path box). The dashed line at $0.5$ applies only to the amber AUROC, not to the teal flip
rate.}
\label{supp:fig:robust}
\end{figure}

\begin{figure}[t]
\centering
\includegraphics[width=\linewidth]{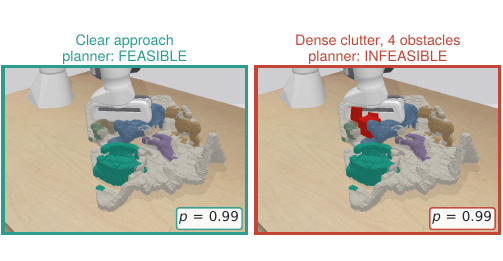}
\caption{Dense multi-object clutter on a GraspNet scene: a cluster of four solid obstacles inserted
into the approach makes the planner relabel the grasp infeasible (right), while the deployed SOTA returns
the same $p(\text{feasible})$ as the clear scene.}
\label{supp:fig:clutter}
\end{figure}

\paragraph{Dense multi-object clutter.}
The diagnosis also survives dense clutter rather than a single inserted body. Inserting a cluster of
four solid obstacles into the approach flips all $150$ test grasps the planner
re-checks (causal flip $1.00$, across two failure modes: $149$ blocked-approach and $1$
blocked-transit), yet the deployed SOTA returns the same $p(\text{feasible})$ with and without the
cluster (CF-pair AUROC $0.498$, $|\Delta p| < 0.001$); \cref{supp:fig:clutter} shows one example.

\begin{figure}[t]
\centering
\includegraphics[width=\linewidth]{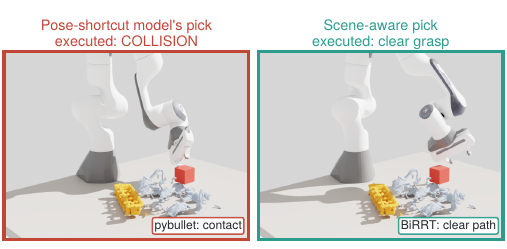}
\caption{Closed-loop safety demo on one counterfactual scene (Novel split, scene $160$), path-traced in
Isaac Sim by replaying the pybullet link poses. \emph{Left}: the pose-shortcut model's top pick
drives the gripper into the inserted box (pybullet-verified collision). \emph{Right}: the clearance-aware screen
of the same $20$ candidates returns a collision-free top-1 (BiRRT-verified).}
\label{supp:fig:blindsafe}
\end{figure}

\paragraph{The result has a physical cost.}
The score gap is not only a ranking artifact; it changes what a robot does. On one counterfactual
scene (Novel split, scene $160$) we execute the candidate grasps in physics simulation
(\cref{supp:fig:blindsafe}). The pose-shortcut model's top-ranked grasp, the one it scores feasible,
drives the gripper into the inserted corridor box, a pybullet-verified contact. Screening the same $20$
candidate grasps with the clearance-aware read of \cref{sec:geometry} instead returns a top-1 whose
approach is collision-free, verified by a BiRRT path. This is one scene, illustrating the score's
downstream effect, not a deployment claim. Same scene, same candidates: the deciding difference is whether the score
reads the geometry the planner checks.

\appendix

\section{The protocol extends to multi-step plan feasibility}
\label{supp:seq}

\begin{figure}[t]
\centering
\includegraphics[width=\linewidth]{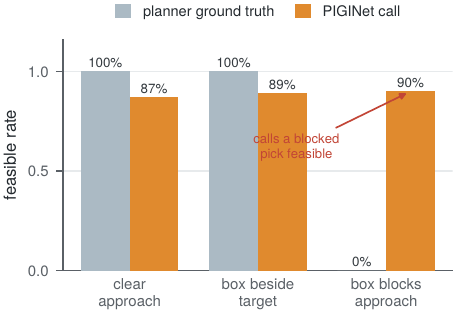}
\caption{PIGINet on its own PR2 kitchen (released weights), aggregated over $450$ counterfactual pairs.
Its binarized feasibility call is flat ($87$ to $90\%$) across a clear approach, a harmless box beside the
target, and a box blocking the approach, while the planner ground truth drops to zero for the blocking
case.}
\label{supp:fig:kitchen_agg}
\end{figure}

\begin{figure}[t]
\centering
\includegraphics[width=\linewidth]{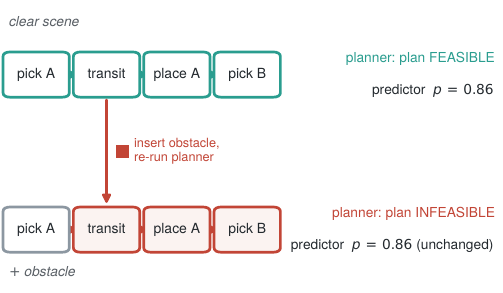}
\caption{The protocol at the plan level. A multi-step rearrangement plan is plan-feasible when the scene
is clear (top); inserting an obstacle at one step makes the planner relabel that step and every step after
it infeasible (bottom), yet a plan-feasibility predictor returns the same score.}
\label{supp:fig:seq}
\end{figure}

Single-grasp feasibility is one action; many planners predict feasibility for a whole action
\emph{sequence}. Our protocol applies unchanged (\cref{supp:fig:seq}): insert an obstacle, re-run the
oracle on the edited state, and relabel each step and the suffix it gates. Two points make this
concrete. \textbf{(i)} PIGINet~\cite{yang2023piginet}, one of the audited models, is itself a
sequence-based plan-feasibility predictor; run from its released weights on its authors' own PR2
kitchen it calls a planner-blocked plan feasible $90\%$ of the time
(\cref{supp:fig:kitchen_agg}; \cref{fig:kitchen}, \cref{sec:hometurf}), so the shortcut is already
demonstrated on a multi-step predictor, not only on single-grasp classifiers. \textbf{(ii)} Our
release includes a whole-skeleton feasibility generator that carries symbolic preconditions and
accumulated geometric state across $k$ steps, so the same insert-and-relabel protocol extends to
sequences with no change to the metric.

\section{Shape and contact resolution on the methods' own data}
\label{supp:scn}

\begin{figure}[t]
\centering
\includegraphics[width=\linewidth]{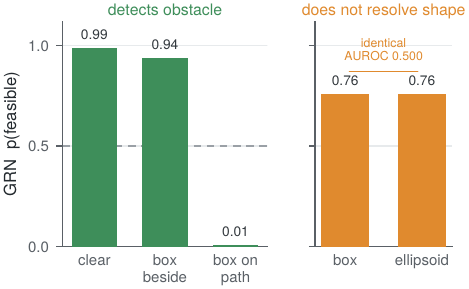}
\caption{GRN on its panda\_3d\_4 benchmark, released weights. Left: GRN detects an inserted obstacle
(feasibility drops only when the box is on the path). Right: a box and its inscribed same-bounding-box
ellipsoid get an identical score (shape-swap AUROC $0.500$).}
\label{supp:fig:grn}
\end{figure}

\cref{fig:hometurf} shows the recreated GRN scene in the main paper; here we give the quantitative
detail of the two released-weights results that are easiest to misread, and the recreated SceneCollisionNet
scenes (\cref{fig:scn}). \textbf{GRN} detects an
inserted obstacle on its panda\_3d\_4 benchmark (feasibility $0.99$ clear, $0.94$ beside, $0.01$
on-path), but a box and its inscribed same-bounding-box ellipsoid receive an identical score, so the
shape-swap AUROC is exactly $0.500$ and shape is decisive for $14\%$ of configurations
(\cref{supp:fig:grn}): GRN reads the bounding box, not the shape. \textbf{SceneCollisionNet}, run from
the released checkpoint on an ACRONYM-style cluttered tabletop in its trained workspace, does not resolve its
own contact boundary: as a gripper-held cube is swept against a solid obstacle from overlapping it to a
$6$\,cm clear gap (\cref{fig:scn}), the released score stays pinned at $p_{\mathrm{coll}}=0.76$
(full-sweep range $0.75$ to $0.77$, never below $0.5$), scoring a deep collision and a $6$\,cm clearance
identically where an ideal checker steps from $1$ to $0$. At its $12.5$\,cm voxel it is geometry-agnostic
below the cell size: in controlled box probes it registers gross presence (removing the obstacle drops
the score $0.67$ to $0.12$) but resolves shape only weakly (same-bounding-box shape swap, $0.821$ AUROC,
not a decision boundary). The common factor across all three released-weights
systems is geometry-agnosticism below each representation's resolution.

\section{Per-model training and architecture}
\label{supp:training}

Every re-implemented predictor is trained on the same GraspNet-1B planner labels under one protocol
(shared oracle, splits, and a binary-cross-entropy objective); the three GraspFC classifiers are the
feasibility predictors of~\cite{sajid2026feasibility}, and GRN is run from released weights on its own data. \cref{supp:tab:cards}
gives each model's representation, training pool, and in-distribution accuracy, so a reader can see that
the interventional gap of \cref{tab:saturation} is not an under-training artifact: every model reaches
high in-distribution AUROC on the same labels before its score falls under the intervention. Class balance
is the benchmark's natural $24.1\%$ feasible on the 100-scene training split (the full label pool is $9.6\%$ feasible); PIGINet's kitchen pool is
$69\%$ feasible, close to the authors' own $67\%$. The three GraspFC classifiers share one recipe
(binary cross-entropy with logits, positive weight $3.14$, AdamW at $1.5{\times}10^{-3}$ with cosine
decay, $30$ epochs, batch $128$) on the benchmark's $100$-scene, $1.52$M-grasp training split; each
re-implementation follows its own paper's loss, noted in \cref{supp:tab:cards}.

\begin{table*}[t]
\centering
\footnotesize
\setlength{\tabcolsep}{4pt}
\begin{tabular}{@{}l p{4.8cm} p{2.0cm} r p{1.4cm} p{2.6cm} r@{}}
\toprule
Model & Representation (what enters the net) & Training pool & Par.\ & Objective & Authors' reported (own data) & Nat.\ \\
\midrule
GraspFC-NNet   & pose $+$ $0.3$\,m grasp-local crop, pooled MLP            & benchmark train split        & 1.65M & BCE & \cite{sajid2026feasibility} & 0.995 \\
GraspFC-Conv3D & pose $+$ voxel grid, 3D CNN                              & benchmark train split        & 9.7M  & BCE & \cite{sajid2026feasibility} & 0.976 \\
GraspFC-PTX    & pose $+$ point cloud, PTv3 backbone                      & benchmark train split        & 9.0M  & BCE & \cite{sajid2026feasibility} & 0.996 \\
SceneCollisionNet & per-point MLP, scatter-max to $12.5$\,cm voxel, 3D conv; PointNet++ gripper & $18$k labels & 26M & hard-neg BCE & acc $93.2\%$, AP $0.990$ \cite{danielczuk2021scenecollision} & 0.964 \\
PIGINet        & per-object CLIP ViT-B/32 image $+$ pose token, $4$ attention blocks & $12$k labels & 31M & pos-weighted BCE & val acc $0.89$, $80\%$ plan-time cut \cite{yang2023piginet} & 0.960 \\
NFC            & object-centered $2$-ch depth $100{\times}100$ (ch.\,2 $=$ neighbors), $2$ conv $+$ $4$ FC & $30$k labels & 22.8k & weighted BCE & acc $93.6\%$, $63.5\%$ time saved \cite{xu2022nfc} & 0.965 \\
AGFP-Net       & $5$ orthographic depth views $+$ target mask, $3{\times}$ ResNet-18 & $18$k labels & 33M & BCE & plan success $90$ to $100\%$ \cite{bouhsain2024agfp} & 0.920 \\
\midrule
GRN (released) & box $[l,w,h,x,y,z,\mathrm{yaw}]$ graph, GNN $+$ supervised obstruction head & panda\_3d\_4 ($20$k scenes), released weights & 2.3M & (released) & F1 $0.94/0.94/0.99$, MAE $0.028$ \cite{bouhsain2025grn} & 0.995 \\
\bottomrule
\end{tabular}
\caption{Per-model representation, training pool, and in-distribution accuracy. \emph{Nat.}\ is the
shared-benchmark natural AUROC of \cref{tab:saturation} (GRN on its own panda\_3d\_4 task); \emph{Authors'
reported} is each method's published headline on its own data, none of which is AUROC, which is why we
re-measure all of them under one Nat. The re-implementations are trained on planner labels under one
protocol; PIGINet loads its released head and the GRN run matches the paper's F1/MAE.}
\label{supp:tab:cards}
\end{table*}

\section{Per-model interventional results and the representation-to-failure taxonomy}
\label{supp:taxonomy}

\cref{tab:saturation} reports natural and CF-pair AUROC; here we read it two ways. First, the
representation-to-failure taxonomy (\cref{supp:tab:taxonomy}) names, for each model, the coarsest
collision geometry its representation resolves and the mechanism behind the gap; the single common
axis is that each model is geometry-agnostic below its representation's resolution. Second, the behavioral
gallery (\cref{supp:tab:behavior}) gives the score each model returns \emph{with and without} the
flip-causing obstacle on a matched set: five predictors return essentially the clear-scene score
($|\Delta p| \le 0.02$) while the planner flips the label, and GRN, the one model with a supervised
obstruction head fed the box, moves with the planner. The obstacle is verified present in every model's
input (it renders into NFC's dedicated neighbor channel, into all five of AGFP-Net's depth views, and, for
the deployed SOTA, a frame audit re-places it on the approach corridor in the model's own input frame;
\cref{supp:neg}), so a flat score is the representation discarding visible evidence, not a missing input.

\begin{table*}[t]
\centering
\small
\setlength{\tabcolsep}{6pt}
\begin{tabular}{@{}l p{2.6cm} r r p{6.5cm}@{}}
\toprule
Model & Resolves at & Nat.\ & CF-pair & Failure mechanism \\
\midrule
GraspFC-NNet      & crop pool       & 0.995 & 0.491 & registers the obstacle; feature displacement is direction-random \\
GraspFC-Conv3D    & voxel grid      & 0.976 & 0.497 & obstacle in the voxel grid, discarded \\
GraspFC-PTX       & point pool      & 0.996 & 0.498 & obstacle in the cloud, discarded \\
SceneCollisionNet & $12.5$\,cm voxel & 0.964 & 0.470 & contact boundary below the $12.5$\,cm voxel \\
PIGINet           & object token    & 0.960 & 0.500 & one object token, no clearance encoded \\
NFC               & $2$-ch depth    & 0.965 & 0.504 & predicts from pose, does not use neighbor channel \\
AGFP-Net          & $5$-view depth  & 0.920 & 0.498 & obstacle in all $5$ views, discarded \\
\midrule
GRN (released)    & box $+$ yaw     & 0.995 & 0.500$^{*}$ & detects box; does not resolve sub-box shape (\cref{supp:grn}) \\
\bottomrule
\end{tabular}
\caption{Representation-to-failure taxonomy. ``Resolves at'' is the coarsest collision geometry each
representation can express; Nat.\ and CF-pair are from \cref{tab:saturation}. $^{*}$GRN moves on presence
(it has a supervised obstruction head fed the box), so its entry is the same-bounding-box shape swap,
dissected in \cref{supp:grn}.}
\label{supp:tab:taxonomy}
\end{table*}

\begin{table}[t]
\centering
\small
\setlength{\tabcolsep}{6pt}
\begin{tabular}{@{}lrrr@{}}
\toprule
Model & clear $p$ & obstacle $p$ & $\Delta p$ \\
\midrule
GraspFC-PTX       & 0.947 & 0.947 & $0.000$ \\
SceneCollisionNet & 0.70  & 0.72  & $+0.02$ \\
PIGINet           & 0.766 & 0.766 & $0.000$ \\
NFC               & 0.798 & 0.797 & $-0.001$ \\
AGFP-Net          & 0.702 & 0.701 & $+0.001$ \\
\midrule
GRN (released)    & 0.990 & 0.007 & $-0.983$ \\
\bottomrule
\end{tabular}
\caption{Behavioral gallery: predicted $p(\text{feasible})$ on a matched set before and after inserting
the flip-causing obstacle, which the planner relabels feasible$\to$infeasible in every row. Values are
means over the matched set; the worked example of \cref{fig:gallery} ($0.975$) is one such pair.}
\label{supp:tab:behavior}
\end{table}

\section{Why GRN partially passes, and where it does not}
\label{supp:grn}

GRN is the one audited model that reacts to an inserted obstacle, so it deserves a careful reading: it is
neither a counterexample to the diagnosis nor a solved case. The difference is structural. GRN has an
explicit, supervised, factored grasp-obstruction head that is fed each object's box and pose, which the
monolithic predictors lack, and on the two things that head can express it is causally faithful.
\textbf{(i) Presence.} Removing the true obstructor flips GRN to feasible $94/91/90\%$ of the time across
$4/10/20$-object scenes, while removing a random non-obstructor flips it $0\%$, ruling out an
any-edge-removal artifact. On the same insertion probe we apply to the other models, GRN scores clear
versus blocked AUROC $1.000$ ($p\,0.990 \to 0.007$; a beside control stays at $0.943$): the same probe
that leaves the other models at chance moves GRN end to end. Instrumenting GRN's graph construction confirms the inserted node enters the
computation and its obstruction head fires proportionally to geometry ($1.00$ in the approach versus
$0.28$ beside), so the discriminator is the head's reasoning, not mere edge presence.
\textbf{(ii) Bounding-box clearance.} Shrinking or sliding the obstacle moves GRN's score smoothly and
monotonically, crossing $0.5$ right as the boxes separate; it tracks clearance magnitude, not just
presence, at the bounding-box level.

Where it fails is below that representation. GRN's per-object input is an axis-aligned box plus yaw, with
no shape and no out-of-plane pose, so any geometry the box cannot encode is invisible by construction, and
the failures span both error directions (\cref{supp:tab:grn}). A solid box and its inscribed
same-bounding-box ellipsoid receive an identical prediction, yet $14\%$ of configurations are
shape-decisive (the box intersects the approach corridor while the ellipsoid does not), giving a shape-swap
AUROC of exactly $0.500$. A center-fixed obstacle tilted into the approach is identical to the same
object standing upright, because pitch and roll are dropped at the node, so on $72.5\%$ of these
tilted-obstacle cases GRN reads the object as upright and commands a colliding grasp: a false-clear, the
unsafe direction. An open frame or bin is filled to its
solid bounding box, so GRN abandons a reachable grasp $100\%$ of the time: a false-block. Each is a matched
pair GRN scores identically while true feasibility flips, so its AUROC is chance on all three against an
oracle's $1.0$.

For a datasets venue, these geometries are out of GRN's box-only benchmark
distribution, and that is the point: GRN's high score on its own benchmark is partly due to the
benchmark's limited geometry (axis-aligned boxes), and the counterfactuals expose the failures the
dataset hides. This is a dataset-and-representation critique, not a claim that GRN is broken. (A practitioner could
give GRN the recomputed bounding box of a tilted object, but that is external geometry GRN does not compute,
and it would then over-block; the native representation assumes upright boxes.)

\begin{table}[t]
\centering
\small
\setlength{\tabcolsep}{6pt}
\begin{tabular}{@{}lrl@{}}
\toprule
Hole & GRN AUROC & Mislabels \\
\midrule
Shape     & 0.500 & $14\%$ (false block) \\
Tilt      & 0.500 & $72.5\%$ (false clear) \\
Concavity & 0.500 & $100\%$ (false block) \\
\bottomrule
\end{tabular}
\caption{GRN passes presence and bounding-box clearance but is at chance (AUROC $0.5$) for geometry
finer than its bounding box: three sub-bounding-box holes (shape, tilt, concavity), each a matched pair
GRN scores identically while the corridor oracle flips. ``Mislabels'' is the fraction of probed targets
the hole gets wrong; the failures span both error directions (false block and false clear).}
\label{supp:tab:grn}
\end{table}

\section{Negative controls and label integrity}
\label{supp:neg}

The diagnosis rests on the flip being caused by the true obstacle and on that obstacle being present in
the model's input, so we report the controls that close the obvious alternative explanations.

\noindent\textbf{The obstacle is on the corridor in the model's own frame.} A concern is that the obstacle
might float off the approach in the model's input frame, which would make $\Delta p \approx 0$ a placement
artifact. A frame audit re-places the box on the corridor in the deployed SOTA's input frame and
re-scores: the model still does not respond (clear-versus-box AUROC $0.498 \to 0.501$, $\Delta p \approx 0$), so
the result is not a misplacement artifact.

\noindent\textbf{Flips are caused by the true obstructor, not any edit.} A matched box placed $30$\,cm
off-corridor leaves the grasp feasible $98.0\%$ of the time, and the same holds for a box slid laterally off
the finger axis ($97\%$ feasible); for GRN, removing a random non-obstructor flips the label $0\%$ of the
time versus $94\%$ for the true obstructor. The edits are causal and specific.

\noindent\textbf{Non-diagnostic perturbations are excluded by design.} A perturbation is only informative if
the oracle relabels it. Naive cloud edits that do not change the label ($50\%$ point dropout, $5$ to
$10$\,mm jitter) move every model's AUROC by $0.000$, which shows only that the model is insensitive to the
cloud, not that it is wrong; we therefore never count them, and use only the insert-and-relabel intervention.

\noindent\textbf{Labels are effectively deterministic.} Re-running the planner with $8$ seeds on held-out
grasps gives $0/480$ flips (label noise $\le 0.6\%$ at $95\%$ confidence), so the $0.99$-level natural scores
are not noise-limited.

\noindent\textbf{The natural label is pose-recoverable, which is why natural data cannot supervise the
term.} A $25$-nearest-neighbor lookup on grasp pose alone, with no scene at all, scores $0.963$ natural AUROC
on SceneCollisionNet's task. The natural distribution carries the feasibility label in the pose, so a model
is never forced to read the scene, which is the standing reason the counterfactual contrast is necessary
rather than convenient.

\noindent\textbf{The in-crop gap is representational, not an information limit.} When the obstacle leaves
the input crop (a distal control $30$\,cm back), the cures fall to chance (raw $0.517$, clearance tokens
$0.505$): an information limit, distinct from the in-crop case where the obstacle is present but discarded.
The global-input cure (from-scratch on counterfactuals, full cloud) partially survives the distal move
($0.602$, versus $0.497$ for the as-is deployed model, \cref{tab:placement}), confirming the in-crop drop
is about the representation, not the input.

\section{The clearance-aware tokenization (architecture and scope)}
\label{supp:ict}

\begin{figure*}[t]
\centering
\includegraphics[width=\linewidth]{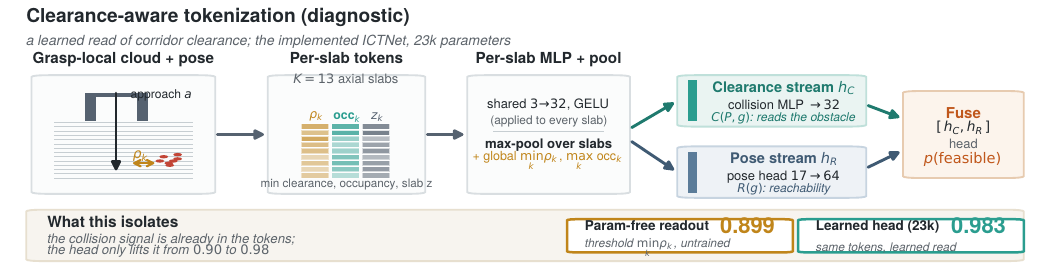}
\caption{The clearance-aware tokenization of \cref{sec:geometry}, the implemented $23$k-parameter
network. The grasp-local cloud and pose enter in the gripper frame; the $10$\,cm approach corridor is
sliced into $K=13$ axial slabs, each emitting a three-number token (minimum lateral clearance $\rho_k$,
occupancy, slab center). A shared per-slab MLP, a max-pool over slabs, and the global $\min_k\rho_k$ form
a clearance feature $h_C$ fused with a pose feature $h_R$ (the $R\otimes C$ factorization).}
\label{fig:ict}
\end{figure*}

\cref{sec:geometry} shows that the collision term, which no audited predictor resolves at the scale that decides the collision, is recoverable
at train time from the original benchmark by a clearance-aware read of the grasp-local points. This
appendix gives the construction behind that result (\cref{fig:ict}) and states what it does and
does not establish. It is a \emph{diagnostic}: it isolates \emph{where} the missing signal lives (the
representation), not a deployable collision predictor. The
distinction matters because, as \cref{sec:fixes} shows, a learned read of these tokens
is still placement-bound, like the from-scratch counterfactual cure.

\paragraph{Tokenization.}
We work in the gripper frame, where collision is exactly SE(3)-invariant (the obstacle's gripper-frame
coordinates are unchanged by any rigid motion of the scene-plus-grasp). We slice the $10$\,cm approach
corridor into $K=13$ axial slabs along the approach axis $a$. Each non-target scene point is projected
onto $a$ to get an axial coordinate $z$ and a lateral radius $\rho$ (its perpendicular distance from the
approach line); points outside the corridor extent are dropped. Slab $k$ emits a three-number token
$[\,\rho_k,\,\mathrm{occ}_k,\,z_k\,]$: $\rho_k$ is the minimum lateral clearance of any obstacle point in
that slab (capped), $\mathrm{occ}_k$ is its occupancy (a soft point count), and $z_k$ is the slab center.
The result is a fixed $K\times 3$ tensor regardless of how many points fall in the corridor, so the
representation is point-count-agnostic and obstacle-localizing by construction: the channel $\rho_k$ is a
direct readout of where the swept gripper comes closest to matter, the very quantity the planner checks.

\paragraph{Network.}
A shared per-slab MLP ($3\!\to\!32$, GELU) maps each token to an embedding; we max-pool over slabs and
append the two global statistics $\min_k \rho_k$ and $\max_k \mathrm{occ}_k$, then a small collision MLP
produces a clearance feature $h_C$. A separate pose head produces $h_R$ from the same $17$-dimensional
pose descriptor the baselines use, and the two are concatenated into the final feasibility head. This is
the $R\otimes C$ factorization of \cref{fig:ict}: $h_R$ carries reachability (pose, not SE(3)-invariant)
and $h_C$ carries collision (gripper-frame, invariant). The whole network is $23$k parameters; the
per-slab MLP is shared across slabs, so the parameter count is independent of $K$ (a sweep over
$K\in\{5,9,13,21,31\}$ leaves it at $22.8$k and within-grasp AUROC at $0.996$ to $1.0$).

\paragraph{Two readouts, one representation.}
The point of the construction is that the collision signal is present in the tokens \emph{before} any
learning. Thresholding the single global statistic $\min_k \rho_k$, with no trained weights and no labels,
already gives CF-pair AUROC $0.899$ and within-grasp $0.911$. The learned head on the identical tokens
reaches CF-pair $0.983$. Learning therefore \emph{sharpens} a signal the representation already exposes; it
does not manufacture it. This is the central contrast: a pose representation cannot be made to read
collision at any accuracy (\cref{tab:encoders}, \cref{fig:geom}), whereas a clearance representation makes
the term readable at $0.90$ with zero fitting. The gap is the representation, not learnability.

\paragraph{Two train-time innovations on the original data.}
Two structural changes sharpen this clearance read on the original benchmark, with no counterfactual
scenes, and are the constructions behind the ``sharpen'' claim of \cref{sec:geometry}.
\textbf{(i) A per-point clearance channel.} Appending each point's corridor clearance to its input
features raises a raw encoder's clearance-probe $R^2$ from $0.05$ to $0.22$; distilling the same signal
into the \emph{pooled} feature instead leaves it at $0.02$, so the clearance must enter before
pooling.
\textbf{(ii) A soft-min over per-region clearances.} Replacing max-pooling with a soft-min over the
per-slab clearances makes the feasibility head monotone in clearance by construction and cuts the
insertion-monotonicity violation rate, the fraction of random point insertions that \emph{raise}
predicted feasibility, from $0.42$ under max-pooling to $0.025$. Both act at train time on the original
splits, so the collision term is sharpened without any counterfactual data.

\paragraph{What it does not establish.}
The learned read is \emph{not} a transferable collision model. Trained on one corridor placement it
transfers across held-out obstacle \emph{shape} and \emph{size} in ranking (sphere, non-convex L, and
$1.5/3/5$\,cm boxes all transfer at or above the in-distribution score), but its absolute calibration
drifts under size shift and its accuracy falls on an unseen on-axis \emph{placement} family ($0.74$,
\cref{sec:ood}), because workspace clearance equals configuration-space clearance only for a point robot.
A transferable solution must learn a calibrated clearance \emph{field} with the structural
guarantees a binary classifier lacks (insertion-monotonicity, size transfer by Minkowski subtraction, and
a certified screening bound); we treat that as the method that follows this dataset paper, not as a claim
of the present work. Likewise, holding the Franka gripper fixed, varying only the finger aperture or the
on-axis approach depth does not move the planner's label (the arm body dominates the swept volume), so a
true cross-gripper test needs a second arm and is out of scope here.

\paragraph{Out-of-distribution transfer of the cures.}
\cref{fig:ood} gives the per-variant breakdown behind \cref{sec:ood}: three counterfactual-cured
models (a generic architecture, a slab-tokenized model, and the clearance-token model), each
trained on box counterfactuals and re-evaluated, planner-relabeled, on a held-out shape (sphere),
placement (axial), and non-convex (L-shape) family. Every cure transfers at or near its in-distribution
score; on the unseen on-axis placement the generic architecture transfers best ($0.90$) and the
clearance one least ($0.74$), both far above chance, while the no-cloud control stays at $0.500$
throughout.

\section{Datasheet}
\label{supp:datasheet}
This datasheet follows Gebru et al.~\cite{gebru2021datasheets}.

\noindent\textbf{Composition.} The counterfactual benchmark pairs each base-feasible grasp with a
planner-verified clear and obstacle-inserted variant at an identical pose, on public GraspNet-1B
scenes. It includes the $1{,}065$ same-pose CF pairs used for CF-pair AUROC, graded obstacle-offset
groups (one grasp with six offsets from blocking to harmless), and the held-out generalization and
robustness sets used here: held-out shape (sphere), non-convex (L-shape), placement (axial), distal
and adjacent placements, obstacle size ($1.5/3/5$\,cm), standoff ($5/10/15$\,cm), and the
resting-pillar and transit-path obstacle types, plus the $150$ dataset-object pairs of
\cref{fig:indist} and the dense-clutter set of \cref{supp:fig:clutter}. Each instance stores the scene
and grasp identifiers, the obstacle geometry and pose, the planner label, and the binding failure mode.

\noindent\textbf{Collection and labels.} Labels come from an RRT-Connect motion planner for a 7-DOF
Franka Panda in PyBullet (IK at the grasp and a pre-grasp standoff, a collision-free linear approach,
and a collision-free path from home). Each label is a fresh planner call on the edited scene (mean
$2.51$\,s; median $2.37$; $n=40$). The label is effectively deterministic: re-running with $8$ seeds
on held-out grasps gives $0/480$ flips (label noise $\le 0.6\%$ at $95\%$ confidence, rule of three).

\noindent\textbf{Splits and intended use.} Evaluation uses the official Seen/Similar/Novel object
splits, with the counterfactual tests reported on Novel objects. Intended use is auditing
obstacle-awareness of feasibility predictors. The protocol is also applied to three prior-work
settings (SceneCollisionNet's ACRONYM, PIGINet's PR2 kitchen, GRN's panda\_3d\_4) to show it is not
tied to our data.

\noindent\textbf{Responsible use, biases, and limitations.} All labels are produced in simulation by a
planner oracle; the benchmark is a diagnostic of obstacle-awareness and is \emph{not} a deployment
safety certificate, and a high score does not certify a model for use on a physical robot. The data
inherit GraspNet-1B's distribution (tabletop RGB-D scenes and its object set) and our single-arm
Franka Panda setup, so coverage of arms, grippers, and environments is limited. By construction the
natural feasibility label is largely recoverable from the grasp
pose alone (\cref{supp:neg}); this is the bias the counterfactual split is designed to expose rather
than a defect to be corrected. Obstacles are geometrically inserted solid bodies validated by the
planner relabel, not by physical settling.

\noindent\textbf{Distribution and licensing.} We release the planner-verified annotations, the
obstacle geometry, the planner configuration, and the relabeling pipeline so others can score new
models without re-running a planner and can regenerate labels for new obstacle families. The underlying
RGB-D scenes remain under the GraspNet-1B license for research use, so we redistribute only the new
annotations and code, not the source images. We release the new planner-verified annotations and this
datasheet under CC BY 4.0 and the relabeling pipeline code under the MIT license. The annotations,
datasheet, and relabeling pipeline are
available at \url{https://github.com/codewithsajid/graspfc-cf}.

\noindent\textbf{Maintenance, versioning, and contact.} We will host the annotations, the pipeline,
and a public leaderboard of interventional scores (\cref{supp:usage}), version the release so scores
remain comparable across updates, and document any errata or added obstacle families in a changelog.

\section{Using the released benchmark}
\label{supp:usage}

The reusable contribution is that a model's obstacle-awareness can be measured without a motion
planner. To audit a new predictor $M$: (1)~load the released same-pose pairs; (2)~for each pair, have
$M$ score the clear cloud and the obstacle-inserted cloud at the \emph{identical} grasp pose, adding
the released obstacle points to $M$'s input; (3)~report CF-pair AUROC over the pairs, within-grasp
AUROC over the graded-offset groups, and pick-clear accuracy. A model that reads only the grasp pose scores chance on
all three by construction: on the released pairs the deployed SOTA and a pose-only baseline both score
pick-clear $0.500$, exactly as on CF-pair, while a clearance-aware read of the same points reaches
$0.997$ (raw corridor points recover part of it, $0.817$), so the metrics cannot be satisfied by a pose
lookup. No planner call is
needed at audit time: scoring the $1{,}065$-pair head-to-head set takes under a minute on one GPU,
versus the $\approx 45$ minutes of planner time the labels themselves cost. Each obstacle family
(shape, placement, size, standoff, resting, transit) is a separate split on which all three metrics are
reported, so a submission can show which interventions a model survives. Regenerating labels for a \emph{new} obstacle family is the
only step that calls the planner, and the released pipeline does it from a single geometry
specification. We will host the annotations, pipeline, and a public leaderboard of interventional
scores.

%% file: main.bbl
\begin{thebibliography}{37}
\providecommand{\natexlab}[1]{#1}
\providecommand{\url}[1]{\texttt{#1}}
\expandafter\ifx\csname urlstyle\endcsname\relax
  \providecommand{\doi}[1]{doi: #1}\else
  \providecommand{\doi}{doi: \begingroup \urlstyle{rm}\Url}\fi

\bibitem[Ait~Bouhsain et~al.(2024)Ait~Bouhsain, Alami, and
  Simeon]{bouhsain2024agfp}
Smail Ait~Bouhsain, Rachid Alami, and Thierry Simeon.
\newblock Extending task and motion planning with feasibility prediction:
  Towards multi-robot manipulation planning of realistic objects.
\newblock In \emph{IEEE/RSJ International Conference on Intelligent Robots and
  Systems (IROS)}, 2024.

\bibitem[Ait~Bouhsain et~al.(2025)Ait~Bouhsain, Alami, and
  Simeon]{bouhsain2025grn}
Smail Ait~Bouhsain, Rachid Alami, and Thierry Simeon.
\newblock Learning geometric reasoning networks for robot task and motion
  planning.
\newblock In \emph{International Conference on Learning Representations
  (ICLR)}, 2025.

\bibitem[Ansari et~al.(2026)Ansari, Arthi, Varma, and
  Thomas]{sajid2026feasibility}
Sajid Ansari, Arthi, Girish Varma, and Antony Thomas.
\newblock Learning motion feasibility from point clouds in cluttered
  environments.
\newblock \emph{arXiv preprint arXiv:2606.26700}, 2026.

\bibitem[Coumans and Bai(2016--2021)]{coumans2021pybullet}
Erwin Coumans and Yunfei Bai.
\newblock {PyBullet}, a {Python} module for physics simulation for games,
  robotics and machine learning.
\newblock \url{http://pybullet.org}, 2016--2021.

\bibitem[Danielczuk et~al.(2021)Danielczuk, Mousavian, Eppner, and
  Fox]{danielczuk2021scenecollision}
Michael Danielczuk, Arsalan Mousavian, Clemens Eppner, and Dieter Fox.
\newblock Object rearrangement using learned implicit collision functions.
\newblock In \emph{IEEE International Conference on Robotics and Automation
  (ICRA)}, 2021.

\bibitem[Dantam et~al.(2018)Dantam, Kingston, Chaudhuri, and
  Kavraki]{dantam2018IJRR}
Neil~T. Dantam, Zachary~K. Kingston, Swarat Chaudhuri, and Lydia~E. Kavraki.
\newblock An incremental constraint-based framework for task and motion
  planning.
\newblock \emph{International Journal of Robotics Research, Special Issue on
  the 2016 Robotics: Science and Systems Conference}, 37\penalty0
  (10):\penalty0 1134--1151, 2018.

\bibitem[de~Haan et~al.(2019)de~Haan, Jayaraman, and Levine]{dehaan2019causal}
Pim de Haan, Dinesh Jayaraman, and Sergey Levine.
\newblock Causal confusion in imitation learning.
\newblock In \emph{Advances in Neural Information Processing Systems
  (NeurIPS)}, 2019.

\bibitem[Deng et~al.(2021)Deng, Litany, Duan, Poulenard, Tagliasacchi, and
  Guibas]{deng2021vectorneurons}
Congyue Deng, Or Litany, Yueqi Duan, Adrien Poulenard, Andrea Tagliasacchi, and
  Leonidas~J. Guibas.
\newblock Vector neurons: A general framework for {SO(3)}-equivariant networks.
\newblock In \emph{Proceedings of the IEEE/CVF International Conference on
  Computer Vision (ICCV)}, 2021.

\bibitem[Emukpere et~al.(2026)Emukpere, Deffayet, and
  Renders]{emukpere2026brittle}
David Emukpere, Romain Deffayet, and Jean-Michel Renders.
\newblock Robust skills, brittle grounding: Diagnosing restricted
  generalization in vision-language action policies via multi-object picking.
\newblock \emph{arXiv preprint arXiv:2602.24143}, 2026.

\bibitem[Eppner et~al.(2021)Eppner, Mousavian, and Fox]{eppner2021acronym}
Clemens Eppner, Arsalan Mousavian, and Dieter Fox.
\newblock {ACRONYM}: A large-scale grasp dataset based on simulation.
\newblock In \emph{IEEE International Conference on Robotics and Automation
  (ICRA)}, 2021.

\bibitem[Fang et~al.(2020)Fang, Wang, Gou, and Lu]{fang2020graspnet}
Hao-Shu Fang, Chenxi Wang, Minghao Gou, and Cewu Lu.
\newblock {GraspNet-1Billion}: A large-scale benchmark for general object
  grasping.
\newblock In \emph{Proceedings of the IEEE/CVF Conference on Computer Vision
  and Pattern Recognition (CVPR)}, 2020.

\bibitem[Gardner et~al.(2020)Gardner, Artzi, Basmova,
  et~al.]{gardner2020contrast}
Matt Gardner, Yoav Artzi, Victoria Basmova, et~al.
\newblock Evaluating models' local decision boundaries via contrast sets.
\newblock In \emph{Findings of the Association for Computational Linguistics:
  EMNLP}, 2020.

\bibitem[Garrett et~al.(2018)Garrett, Lozano-P{\'e}rez, and
  Kaelbling]{garrett2018IJRR}
Caelan~Reed Garrett, Tom{\'a}s Lozano-P{\'e}rez, and Leslie~Pack Kaelbling.
\newblock {FFRob}: Leveraging symbolic planning for efficient task and motion
  planning.
\newblock \emph{The International Journal of Robotics Research}, 37\penalty0
  (1):\penalty0 104--136, 2018.

\bibitem[Gebru et~al.(2021)Gebru, Morgenstern, Vecchione, Vaughan, Wallach,
  Daum{\'e}~III, and Crawford]{gebru2021datasheets}
Timnit Gebru, Jamie Morgenstern, Briana Vecchione, Jennifer~Wortman Vaughan,
  Hanna Wallach, Hal Daum{\'e}~III, and Kate Crawford.
\newblock Datasheets for datasets.
\newblock \emph{Communications of the ACM}, 64\penalty0 (12), 2021.

\bibitem[Geirhos et~al.(2019)Geirhos, Rubisch, Michaelis, Bethge, Wichmann, and
  Brendel]{geirhos2019texture}
Robert Geirhos, Patricia Rubisch, Claudio Michaelis, Matthias Bethge, Felix~A.
  Wichmann, and Wieland Brendel.
\newblock {ImageNet}-trained {CNN}s are biased towards texture; increasing
  shape bias improves accuracy and robustness.
\newblock In \emph{International Conference on Learning Representations
  (ICLR)}, 2019.

\bibitem[Geirhos et~al.(2020)Geirhos, Jacobsen, Michaelis, Zemel, Brendel,
  Bethge, and Wichmann]{geirhos2020shortcut}
Robert Geirhos, J{\"o}rn-Henrik Jacobsen, Claudio Michaelis, Richard Zemel,
  Wieland Brendel, Matthias Bethge, and Felix~A. Wichmann.
\newblock Shortcut learning in deep neural networks.
\newblock \emph{Nature Machine Intelligence}, 2, 2020.

\bibitem[Kaelbling and Lozano-P{\'e}rez(2013)]{kaelbling2013IJRR}
Leslie~Pack Kaelbling and Tom{\'a}s Lozano-P{\'e}rez.
\newblock Integrated task and motion planning in belief space.
\newblock \emph{The International Journal of Robotics Research}, 32\penalty0
  (9-10):\penalty0 1194--1227, 2013.

\bibitem[Karaman and Frazzoli(2011)]{karaman2011IJRR}
Sertac Karaman and Emilio Frazzoli.
\newblock Sampling-based algorithms for optimal motion planning.
\newblock \emph{The International Journal of Robotics Research}, 30\penalty0
  (7):\penalty0 846--894, 2011.

\bibitem[Kaushik et~al.(2020)Kaushik, Hovy, and Lipton]{kaushik2020cad}
Divyansh Kaushik, Eduard Hovy, and Zachary~C. Lipton.
\newblock Learning the difference that makes a difference with
  counterfactually-augmented data.
\newblock In \emph{International Conference on Learning Representations
  (ICLR)}, 2020.

\bibitem[Kew et~al.(2020)Kew, Ichter, Bandari, Lee, and
  Faust]{kew2020clearancenet}
J.~Chase Kew, Brian Ichter, Maryam Bandari, Tsang-Wei~Edward Lee, and
  Aleksandra Faust.
\newblock Neural collision clearance estimator for batched motion planning.
\newblock In \emph{International Workshop on the Algorithmic Foundations of
  Robotics (WAFR)}, 2020.

\bibitem[Kuffner and LaValle(2000)]{kuffner2000rrtconnect}
James~J. Kuffner and Steven~M. LaValle.
\newblock {RRT-Connect}: An efficient approach to single-query path planning.
\newblock In \emph{IEEE International Conference on Robotics and Automation
  (ICRA)}, 2000.

\bibitem[McCoy et~al.(2019)McCoy, Pavlick, and Linzen]{mccoy2019hans}
R.~Thomas McCoy, Ellie Pavlick, and Tal Linzen.
\newblock Right for the wrong reasons: Diagnosing syntactic heuristics in
  natural language inference.
\newblock In \emph{Proceedings of the Annual Meeting of the Association for
  Computational Linguistics (ACL)}, 2019.

\bibitem[Murali et~al.(2023)Murali, Mousavian, Eppner, Fishman, and
  Fox]{murali2023cabinet}
Adithyavairavan Murali, Arsalan Mousavian, Clemens Eppner, Adam Fishman, and
  Dieter Fox.
\newblock {CabiNet}: Scaling neural collision detection for object
  rearrangement with procedural scene generation.
\newblock In \emph{IEEE International Conference on Robotics and Automation
  (ICRA)}, 2023.

\bibitem[Newbury et~al.(2023)Newbury, Gu, Chumbley, Mousavian, Eppner, Leitner,
  Bohg, Morales, Asfour, Kragic, Fox, and Cosgun]{newbury2023grasp}
Rhys Newbury, Morris Gu, Lachlan Chumbley, Arsalan Mousavian, Clemens Eppner,
  J{\"u}rgen Leitner, Jeannette Bohg, Antonio Morales, Tamim Asfour, Danica
  Kragic, Dieter Fox, and Akansel Cosgun.
\newblock Deep learning approaches to grasp synthesis: A review.
\newblock \emph{IEEE Transactions on Robotics (T-RO)}, 2023.

\bibitem[Qi et~al.(2017{\natexlab{a}})Qi, Su, Mo, and Guibas]{qi2017pointnet}
Charles~R. Qi, Hao Su, Kaichun Mo, and Leonidas~J. Guibas.
\newblock {PointNet}: Deep learning on point sets for {3D} classification and
  segmentation.
\newblock In \emph{Proceedings of the IEEE Conference on Computer Vision and
  Pattern Recognition (CVPR)}, 2017{\natexlab{a}}.

\bibitem[Qi et~al.(2017{\natexlab{b}})Qi, Yi, Su, and Guibas]{qi2017pointnetpp}
Charles~R. Qi, Li Yi, Hao Su, and Leonidas~J. Guibas.
\newblock {PointNet++}: Deep hierarchical feature learning on point sets in a
  metric space.
\newblock In \emph{Advances in Neural Information Processing Systems
  (NeurIPS)}, 2017{\natexlab{b}}.

\bibitem[Qian et~al.(2022)Qian, Li, Peng, Mai, Hammoud, Elhoseiny, and
  Ghanem]{qian2022pointnext}
Guocheng Qian, Yuchen Li, Houwen Peng, Jinjie Mai, Hasan Abed Al~Kader Hammoud,
  Mohamed Elhoseiny, and Bernard Ghanem.
\newblock {PointNeXt}: Revisiting {PointNet++} with improved training and
  scaling strategies.
\newblock In \emph{Advances in Neural Information Processing Systems
  (NeurIPS)}, 2022.

\bibitem[Sundermeyer et~al.(2021)Sundermeyer, Mousavian, Triebel, and
  Fox]{sundermeyer2021contactgrasp}
Martin Sundermeyer, Arsalan Mousavian, Rudolph Triebel, and Dieter Fox.
\newblock {Contact-GraspNet}: Efficient {6-DoF} grasp generation in cluttered
  scenes.
\newblock In \emph{IEEE International Conference on Robotics and Automation
  (ICRA)}, 2021.

\bibitem[Thomas et~al.(2021)Thomas, Mastrogiovanni, and
  Baglietto]{thomas2021RAS}
Antony Thomas, Fulvio Mastrogiovanni, and Marco Baglietto.
\newblock {MPTP}: Motion-planning-aware task planning for navigation in belief
  space.
\newblock \emph{Robotics and Autonomous Systems}, 141:\penalty0 103786, 2021.

\bibitem[Wu et~al.(2024)Wu, Jiang, Wang, Liu, Liu, Qiao, Ouyang, He, and
  Zhao]{wu2024ptv3}
Xiaoyang Wu, Li Jiang, Peng-Shuai Wang, Zhijian Liu, Xihui Liu, Yu Qiao, Wanli
  Ouyang, Tong He, and Hengshuang Zhao.
\newblock Point transformer {V3}: Simpler, faster, stronger.
\newblock In \emph{Proceedings of the IEEE/CVF Conference on Computer Vision
  and Pattern Recognition (CVPR)}, 2024.

\bibitem[Wu et~al.(2025)Wu, DeTone, Frost, Shen, Xie, Yang, Engel, Newcombe,
  Zhao, and Straub]{wu2025sonata}
Xiaoyang Wu, Daniel DeTone, Duncan Frost, Tianwei Shen, Chris Xie, Nan Yang,
  Jakob Engel, Richard Newcombe, Hengshuang Zhao, and Julian Straub.
\newblock Sonata: Self-supervised learning of reliable point representations.
\newblock In \emph{Proceedings of the IEEE/CVF Conference on Computer Vision
  and Pattern Recognition (CVPR)}, 2025.

\bibitem[Wu et~al.(2015)Wu, Song, Khosla, Yu, Zhang, Tang, and
  Xiao]{wu2015modelnet}
Zhirong Wu, Shuran Song, Aditya Khosla, Fisher Yu, Linguang Zhang, Xiaoou Tang,
  and Jianxiong Xiao.
\newblock {3D} {ShapeNets}: A deep representation for volumetric shapes.
\newblock In \emph{Proceedings of the IEEE Conference on Computer Vision and
  Pattern Recognition (CVPR)}, 2015.

\bibitem[Xing et~al.(2025)Xing, Luo, Xie, Gao, Shen, and
  Song]{xing2025shortcut}
Youguang Xing, Xu Luo, Junlin Xie, Lianli Gao, Hengtao Shen, and Jingkuan Song.
\newblock Shortcut learning in generalist robot policies: The role of dataset
  diversity and fragmentation.
\newblock In \emph{Conference on Robot Learning (CoRL)}, 2025.

\bibitem[Xu et~al.(2022)Xu, Ren, Chalvatzaki, and Peters]{xu2022nfc}
Lei Xu, Tianyu Ren, Georgia Chalvatzaki, and Jan Peters.
\newblock Accelerating integrated task and motion planning with neural
  feasibility checking.
\newblock \emph{arXiv preprint arXiv:2203.10568}, 2022.

\bibitem[Yang et~al.(2023)Yang, Garrett, Lozano-P{\'e}rez, Kaelbling, and
  Fox]{yang2023piginet}
Zhutian Yang, Caelan~R. Garrett, Tom{\'a}s Lozano-P{\'e}rez, Leslie Kaelbling,
  and Dieter Fox.
\newblock Sequence-based plan feasibility prediction for efficient task and
  motion planning.
\newblock In \emph{Robotics: Science and Systems (RSS)}, 2023.

\bibitem[Yu and Gao(2021)]{yu2021gnncollision}
Chenning Yu and Sicun Gao.
\newblock Reducing collision checking for sampling-based motion planning using
  graph neural networks.
\newblock In \emph{Advances in Neural Information Processing Systems
  (NeurIPS)}, 2021.

\bibitem[Yu et~al.(2022)Yu, Tang, Rao, Huang, Zhou, and Lu]{yu2022pointbert}
Xumin Yu, Lulu Tang, Yongming Rao, Tiejun Huang, Jie Zhou, and Jiwen Lu.
\newblock {Point-BERT}: Pre-training {3D} point cloud transformers with masked
  point modeling.
\newblock In \emph{Proceedings of the IEEE/CVF Conference on Computer Vision
  and Pattern Recognition (CVPR)}, 2022.

\end{thebibliography}
